\documentclass[letterpaper, 10 pt, conference]{ieeeconf}  %

\IEEEoverridecommandlockouts                 %

\usepackage{graphics} %
\usepackage{epsfig} %
\usepackage{mathptmx} %
\usepackage{times} %
\usepackage{amsmath} %
\usepackage{amssymb}  %
\usepackage{capt-of} %
\usepackage{wrapfig}
\usepackage{tabularx}
 \usepackage{booktabs} 
 \usepackage{xcolor}
\usepackage{url}
\usepackage{gensymb}
\usepackage[font=small, labelfont=bf]{caption}
\usepackage{subcaption}
\usepackage[colorlinks=true,
            linkcolor=blue,
            urlcolor=blue,
            citecolor=blue]{hyperref}
\usepackage[table]{xcolor}

\definecolor{vamos_red}{HTML}{F45D4C}
\definecolor{vamos_blue}{HTML}{4EC5B6}

\title{\LARGE \bf
VAMOS: A Hierarchical Vision-Language-Action Model for Capability-Modulated and
Steerable Navigation
}

\author{Mateo Guaman Castro$^{1}$, Sidharth Rajagopal$^{1}$, Daniel Gorbatov$^{1}$, \\Matt Schmittle$^{1}$, Rohan Baijal$^{1}$, Octi Zhang$^{1}$, Rosario Scalise$^{1}$, Sidharth Talia$^{1}$, \\Emma Romig$^{1}$, Celso de Melo$^{2}$, Byron Boots$^{1}$, Abhishek Gupta$^{1}$ \\$^{1}$ University of Washington, $^{2}$ DEVCOM ARL
}

\begin{document}

\newcommand{\methodname}{\textsc{Vamos}}

\makeatletter
\let\@oldmaketitle\@maketitle%
\renewcommand{\@maketitle}{%
    \@oldmaketitle%
    \centering
    \vspace{1mm}
    \includegraphics[width=\linewidth]{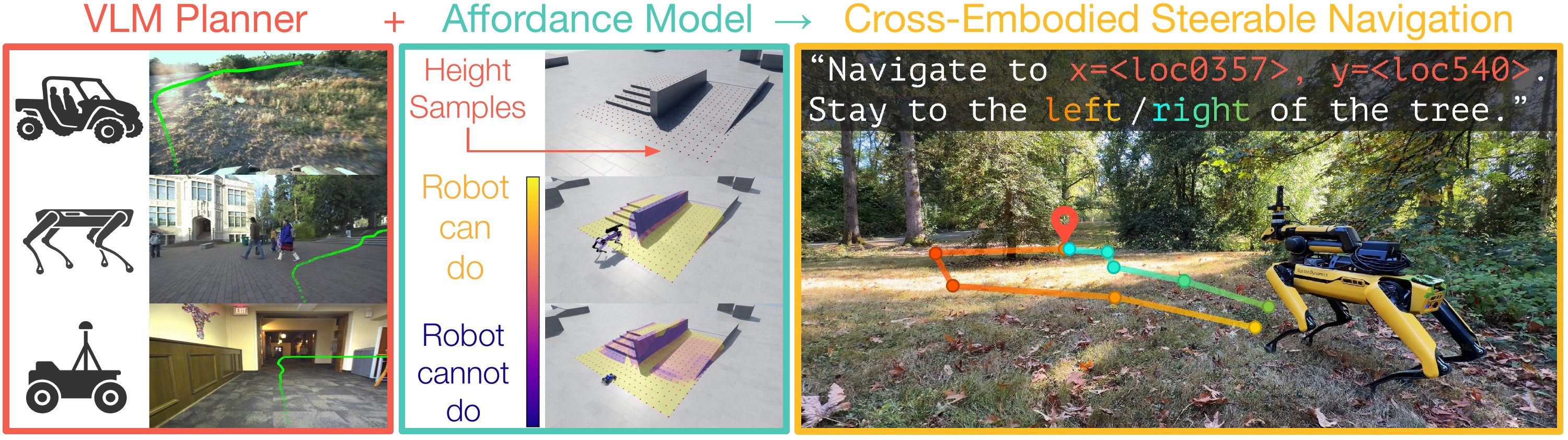}
    \captionof{figure}{We present \textbf{\methodname}, a general-purpose hierarchical VLA for navigation. Our key idea is to decouple semantic planning from embodiment grounding. We achieve this by training a high-level VLM planner with diverse, heterogeneous real-world data that proposes trajectory candidates as 2D paths, which are then re-ranked by an embodiment-specific affordance model trained cheaply and safely in simulation. This yields robust, cross-embodied and steerable open-world navigation controllers.}
    \label{fig:teaser}
    \setcounter{figure}{1}
    \vspace{-0.4cm}
}
\makeatother

\maketitle
\thispagestyle{empty}
\pagestyle{empty}

\begin{abstract}
 A fundamental challenge in robot navigation lies in learning policies that generalize across diverse environments while conforming to the unique physical constraints and capabilities of a specific embodiment (e.g., quadrupeds can walk up stairs, but rovers cannot). We propose \methodname, a hierarchical VLA that decouples semantic planning from embodiment grounding: a generalist planner learns from diverse, open-world data, while a specialist affordance model learns the robot's physical constraints and capabilities in safe, low-cost simulation. We enabled this separation 
by carefully designing an interface that lets a high-level planner propose candidate paths directly in image space that the affordance model then evaluates and re-ranks. Our real-world experiments show that \methodname~achieves higher success rates in both indoor and complex outdoor navigation than state-of-the-art model-based and end-to-end learning methods. We also show that our hierarchical design enables cross-embodied navigation across legged and wheeled robots and is easily steerable using natural language. Real-world ablations confirm that the specialist model is key to embodiment grounding, enabling a single high-level planner to be deployed across physically distinct wheeled and legged robots. Finally, this model significantly enhances single-robot reliability, achieving 3$\times$ higher success rates by rejecting physically infeasible plans. Website: \url{https://vamos-vla.github.io/}

\end{abstract}

\section{Introduction}
\label{sec:introduction}
\begin{figure*}[!h]
    \centering
    \includegraphics[width=1.0\linewidth]{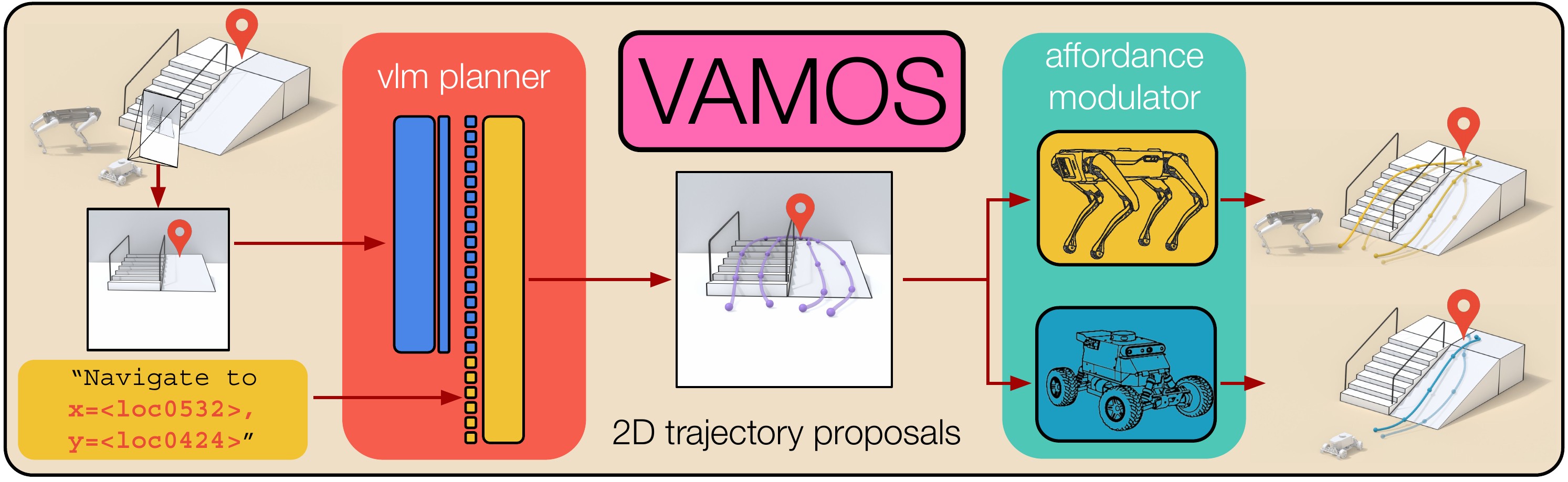}
    \caption{The high-level planner is a VLM trained to take an image and a goal coordinate (encoded as text) as input, optionally appending natural language preferences, and to output a set of candidate paths in pixel space. These paths are encoded as strings of location token pairs, then decoded and projected from 2D pixel space to the 3D ground plane. Finally, a capability-aware affordance function evaluates and re-ranks the 3D candidate paths to determine which path the robot should execute in the real world based on low-level policy capabilities.}
    \label{fig:system_overview}
    \vspace{-5mm}
\end{figure*}
A core problem in robotics is determining how robots can navigate to a goal location while traversing non-trivial terrain and obstacles. The promise of general-purpose robot navigation--- i.e., performing well across diverse environments, different embodiments, and being easy to steer during deployment---has motivated a shift from hand-designed modular stacks to learning-based approaches that leverage large-scale data. Recent advances in robotic foundation models have shown that performance scales with the amount of diverse data provided~\cite{black2024pi0, o2024open, sridhar2024nomad, shah2023vint}. However, as datasets scale, so does their heterogeneity. This becomes a critical challenge when a downstream robot is physically incapable of achieving the entirety of behaviors recorded in a pooled, multi-robot dataset. For instance, data from a quadruped navigating stairs is of limited use to a wheeled robot. This creates a bottleneck that prevents us from naively combining all available data and achieving reliable navigation performance. In this work, we tackle the problem of effectively leveraging large-scale, combined datasets of heterogeneous locomotion capabilities for learning general-purpose cross-embodiment and steerable navigation policies.

To this end, we propose \methodname, a hierarchical vision-language-action (VLA) model. Our key insight is that navigation can be decomposed: high-level heuristics (e.g., reaching a goal, avoiding large obstacles) are generalizable across embodiments, while low-level traversability is strictly dependent on the robot's physical capabilities. \methodname~operationalizes this insight with two main components, i.e., a \textit{high-capacity vision-language model (VLM)} that acts as a generalist high-level planner, and a \textit{lightweight, per-embodiment affordance model} that evaluates the feasibility of the planner's proposed actions. We train the VLM planner on diverse, real-world datasets to instill broad semantic understanding, and we train each embodiment's affordance model in simulation for efficiency and safety (Fig. \ref{fig:teaser}). The interface between these models is a predicted 2D path. This path provides a structured yet flexible representation that enables our planner to leverage heterogeneous data while allowing the affordance model to modulate plans based on embodiment-specific constraints.

Through extensive real-world experiments, we demonstrate that our hierarchical approach, \methodname, yields a new state-of-the-art in general-purpose robot navigation. We show for the first time that a structured VLA can outperform both heavily tuned modular stacks and monolithic foundation models on challenging indoor and outdoor courses. The key to this superior performance is the hierarchical design choices that successfully disentangle general planning from specific physical affordances to  enable cross-embodiment transfer: we achieve high performance on both wheeled and legged robots by reusing the same high-level planner and swapping only a lightweight, specialized affordance model. Our use of a VLM also permits intuitive, natural language steerability at test time. Further, our ablations validate our core design choices, confirming that training with heterogeneous data provides significant positive transfer and that our affordance model is crucial for robust navigation.

\section{Related Work}
\label{sec:related_work}

Our work builds upon three key areas of research: classical modular navigation, end-to-end learning for navigation, and hierarchical vision-language models.

\textbf{Classical Modular Navigation.} Navigation has traditionally been approached using  modular systems with distinct components, e.g., state-estimation, perception, planning, and control~\cite{thorpe1988vision, urmson2008autonomous}. These methods have become the established standard in complex real-world systems due to their reliability and interpretability~\cite{meng2023terrainnet, tranzatto2022cerberus}. To improve their generalization, recent efforts have incorporated learning-based components, e.g., in perception~\cite{shaban2022semantic, erni2023mem}, traversability estimation~\cite{castro2023does, mattamala2024wild, frey2022locomotion, roth2025learned}, or planning~\cite{roth2024viplanner, schmittle2025longrangenavigatorlrn}. 

However, traditional modularity introduces significant limitations. First, these systems are typically heavily tuned for a specific robot embodiment and a bounded set of operating scenarios, making them brittle when deployed in new environments. Second, the intermediate representations, such as 2.5D costmaps, can abstract away valuable information and create performance bottlenecks between modules. Most importantly for our work, these systems lack cross-embodiment generalizability; transferring them to a new robot often requires re-training learned components and extensive re-tuning of the entire stack~\cite{castro2023does, schmittle2025longrangenavigatorlrn}. Our work aims to achieve the robustness of these systems while overcoming their reliance on hand-tuning and their inability to generalize across embodiments.

\textbf{End-to-End Learned Navigation and Foundation Models.} To address the limitations of modular stacks, a dominant paradigm in recent years has been end-to-end learned navigation. This approach seeks to learn a direct mapping from sensor inputs to control actions, shifting the burden from manual system design to large-scale data provision. The success of foundation models in other domains has inspired similar efforts in robotics~\cite{black2024pi0, o2024open, sridhar2024nomad, shah2023vint, rt22023arxiv}, which have demonstrated that policy performance scales effectively with the size and diversity of the training dataset. However, without any additional structure, these methods can be brittle during real-world deployment, e.g., they
often struggle to train across widely heterogeneous datasets due to individual dataset variations in the action space.

\textbf{Hierarchical Architectures and Vision-Language Models.} To achieve a better balance, our work builds upon the paradigm of hierarchical models, which separate high-level planning from low-level control, the latter of which is often treated as an open-loop black box. This structure is well-established in both manipulation~\cite{li2025hamster, gu24rttraj} and navigation~\cite{shah2022gnm, shah2023vint, sridhar2024nomad}. However, the choice of representation and the division of responsibility between the modules are critical. As our experiments later demonstrate, many prior hierarchical models underperform even traditional modular baselines in complex settings. Bidirectional influence between the VLM planner and the affordance module is necessary for robust performance.

One line of work~\cite{shah2022gnm, shah2023vint, sridhar2024nomad} uses a generalist model that takes a goal image as input and outputs a sequence of low-level velocity commands. This approach places an immense burden on a single model to both  learn high-level navigation semantics and infer the specific low-level capabilities of the robot directly from observations. This conflation of tasks compromises performance on anything beyond simple, flat terrain. Moreover, it introduces a practical limitation by requiring a prior demonstration to obtain the goal image and often relies on a pre-built map for long-range navigation, limiting its applicability in unseen environments.

More recently, these hierarchical systems have been instantiated as VLAs, leveraging the semantic reasoning of pre-trained VLMs \cite{cheng2024navila, li2025hamster, glossop2025cast}. The method most relevant to ours is NaVILA~\cite{cheng2024navila}, which finetunes a VLM to map a natural language command to a sequence of textual low-level actions (e.g., "Move forward 25 cm"). This approach has two key drawbacks. First, specifying precise goals via text can be tedious and ambiguous for non-object-centric navigation. Second, discrete, short-horizon textual output commands are not well-suited for long-range planning and, crucially, do not provide a natural interface for downstream modulation by an embodiment-aware module.

We designed \methodname~to overcome these limitations. By predicting a continuous 2D path as our interface, we (1) enable precise, long-range spatial reasoning, (2) do not require prior demonstrations or maps, and (3) create a representation that can be explicitly modulated by our per-embodiment affordance model. This lets our high-level planner focus solely on generalizable navigation strategy, while the affordance model assumes sole responsibility for grounding the plan in the specific robot's physical capabilities.

\section{\methodname: \underline{V}LA for Hierarchical Navigation, \underline{A}ffordance-\underline{Mo}dulated and \underline{S}teerable}
\label{sec:overview}

We propose a learning-based navigation algorithm, \methodname, that can learn from large, heterogeneous datasets while maintaining awareness of embodiment-specific capabilities. To do this, we combine a high-level VLM planner with embodiment-specific, low-level locomotion affordance models, which re-rank the high-level predictions to align with robot capabilities at test time (Fig. \ref{fig:system_overview}). In the following subsections, we outline our high-level generalist model architecture and training paradigm (Section~\ref{sec:data_recipe}) and describe the low-level affordance modulation (Section~\ref{sec:valuemodulation}). 
    
\subsection{High-Level VLM Planners from Large-Scale Datasets}
\label{sec:data_recipe}

A high-level generalist navigation model must be able to incorporate a variety of large-scale data sources, benefiting from their union. To this end, we build on recent advances in vision-language modeling by parameterizing our high-level generalist navigation model as a VLM. Our key design decision then became: \emph{What choice of interface between the high- and low-level models facilitates generic training across heterogeneous datasets
while effectively interfacing with embodiment-specific, low-level control?}

We \textbf{cast high-level navigation as a trajectory prediction problem}, leveraging 2D point prediction as a unifying interface for general-purpose navigation. Specifically, we train a VLM planner $P_\phi(\tau|I, g_l)$ to go from a monocular RGB image $I \in \mathcal{I}$ and target goal coordinates encoded in text $g_l$ to predict a coarse 2D path $\tau \in \mathcal{T}$ in pixel space. The 2D path $\tau$ is a sequence of points that describes a trajectory of where the robot should move in future time-steps, projected onto the image plane for simplicity. Formally, the 2D path is defined as $\tau: (x, y)_{t}$, where $(x, y)$ are normalized pixel locations of the robot's position in the frame at step $t$.  

Our choice of parameterization has several advantages. First, it facilitates general-purpose training from a variety of data sources, with variable action spaces, unified via point prediction. Second, as noted in prior work~\cite{li2025hamster, yuan2024robopoint}, training on point-level predictions helps VLMs retain much of their pre-trained generalization capabilities. The high-level VLM navigation module interfaces with a \emph{low-level} controller $\pi$\textit{ bidirectionally} (see Section~\ref{sec:valuemodulation}); it provides waypoints for the low-level controller to track, while the low-level controller modulates the high-level predictions via its affordance function $F_\pi$.

\begin{figure}[t]
    \centering
    \begin{subfigure}[b]{0.23\columnwidth}
        \centering
        \includegraphics[width=\linewidth]{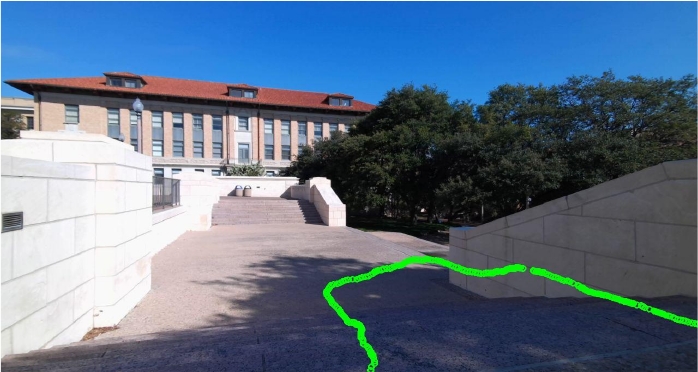}\\
        \includegraphics[width=\linewidth]{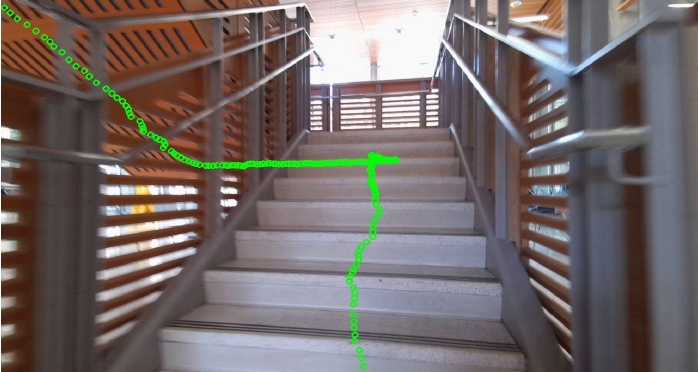}\\
        \includegraphics[width=\linewidth]{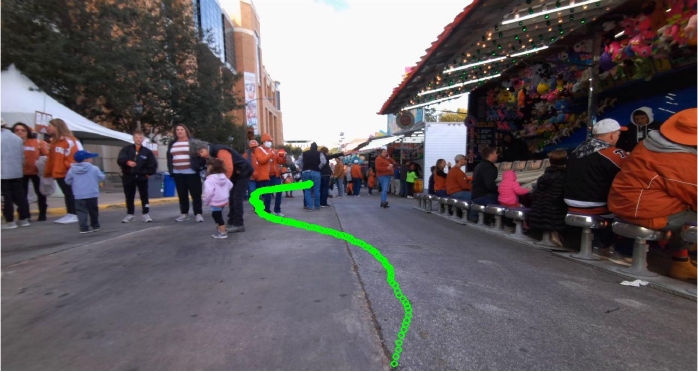}
        \caption{\footnotesize SCAND}
        \label{fig:scand}
    \end{subfigure}
    \hfill
    \begin{subfigure}[b]{0.23\columnwidth}
        \centering
        \includegraphics[width=\linewidth]{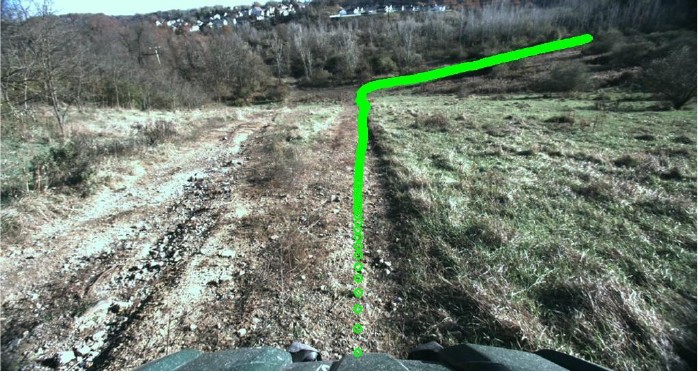}\\
        \includegraphics[width=\linewidth]{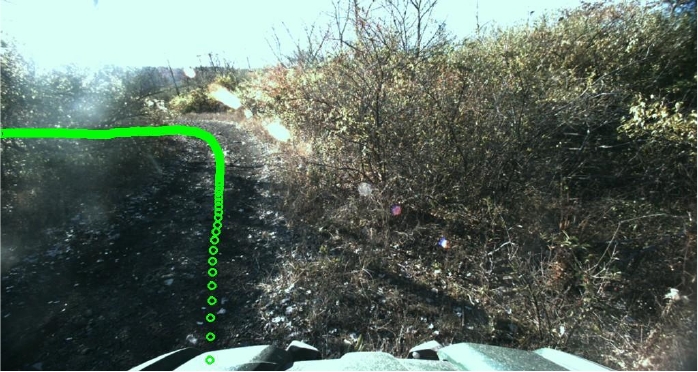}\\
        \includegraphics[width=\linewidth]{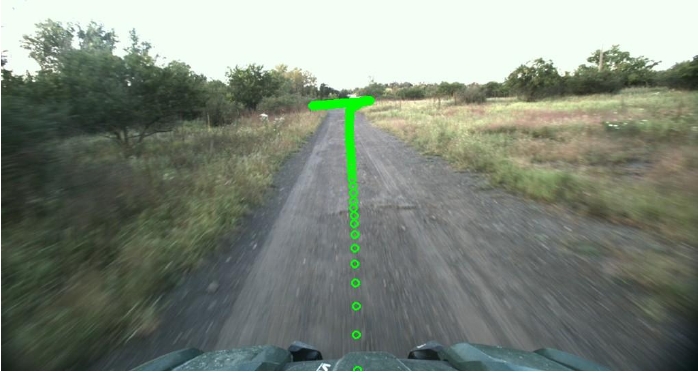}
        \caption{\footnotesize TartanDrive 2}
        \label{fig:tartandrive}
    \end{subfigure}
    \hfill
    \begin{subfigure}[b]{0.23\columnwidth}
        \centering
        \includegraphics[width=\linewidth]{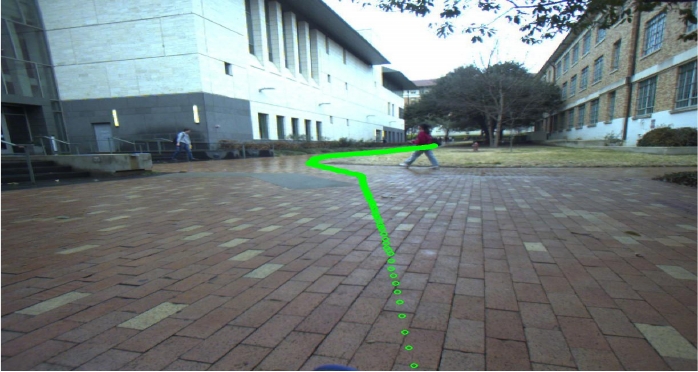}\\
        \includegraphics[width=\linewidth]{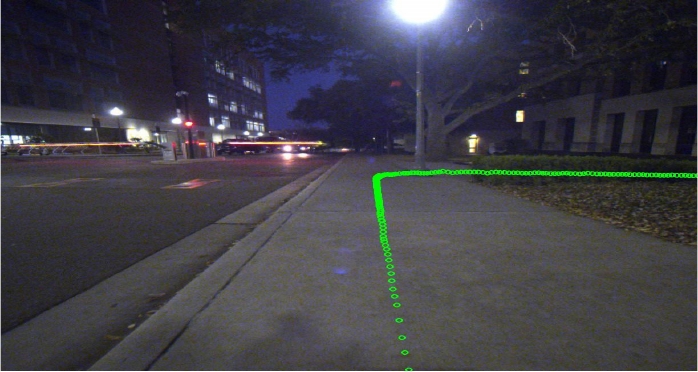}\\
        \includegraphics[width=\linewidth]{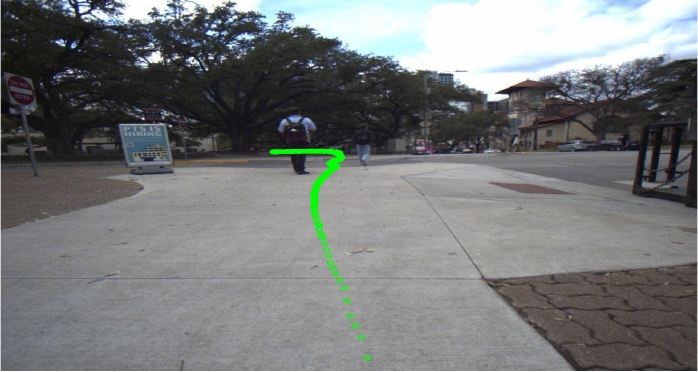}
        \caption{\footnotesize CODa}
        \label{fig:coda}
    \end{subfigure}
    \hfill
    \begin{subfigure}[b]{0.23\columnwidth}
        \centering
        \includegraphics[width=\linewidth]{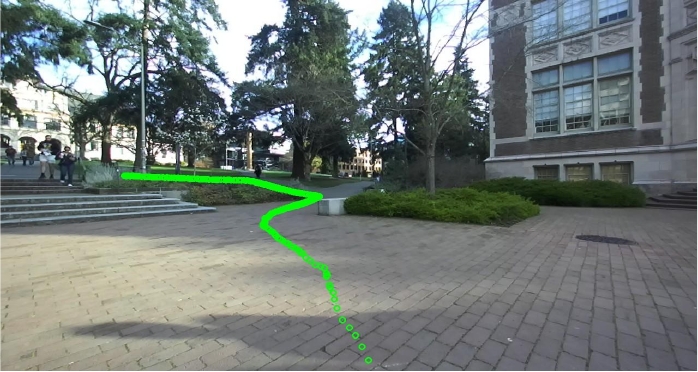}\\
        \includegraphics[width=\linewidth]{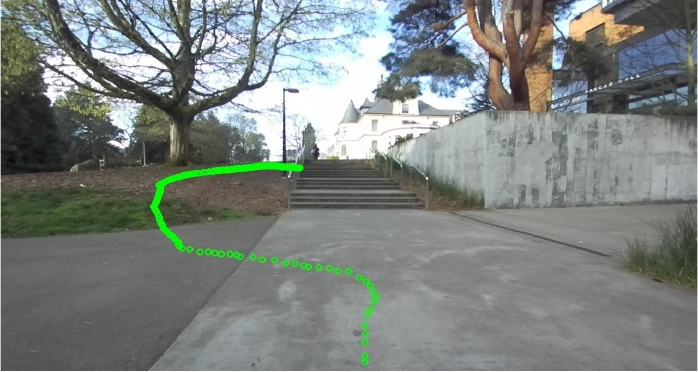}\\
        \includegraphics[width=\linewidth]{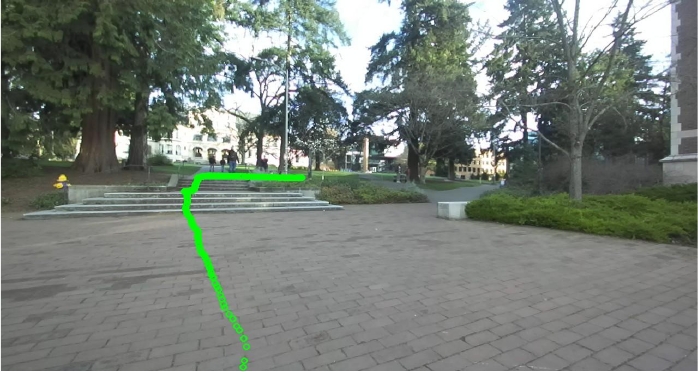}
        \caption{\footnotesize Spot}
        \label{fig:spot_dataset}
    \end{subfigure}

    \caption{We fine-tune a VLM with navigation-specific, real-world datasets, with heterogeneous embodiments and capabilities, to obtain a general-puropse high-level planner. We use a filtered data mix from SCAND \cite{karnan2022socially}, TartanDrive 2 \cite{sivaprakasam2024tartandrive}, CODa \cite{zhang2024towards}, and a small, 0.3 hour in-domain dataset collected on Spot. }
    \label{fig:dataset_depiction}
    \vspace{-7mm}
\end{figure}

To train our steerable VLM planner, we first assemble a diverse navigation dataset mix that spans 29.8 hours and contains odometry-labeled data from 4 different robotic navigation datasets taken from 3 different embodiments (Fig. \ref{fig:dataset_depiction}). We perform a series of data processing and filtering operations (Section \ref{sec:dataset_processing}) that let us obtain higher-quality data for training our navigation generalist. From this dataset, we easily extract labeled data in the form of tuples of images and corresponding navigation paths, represented as 2D points in pixel space. We additionally annotate and augment this data with text descriptions from a state-of-the-art VLM to improve model steerability. 

 Given this training data, we finetune high-level VLMs to perform path predictions given input images and target goal coordinates. We perform supervised finetuning over a pre-trained \texttt{PaliGemma 2 3B} model at $224\texttt{px}^2$ resolution \cite{steiner2024paligemma}.  We use low-rank adapters (LoRAs) since training our models using full-parameter fine-tuning vs LoRA \cite{hu2022lora} yields similar performance.

\subsection{Training Data and Preprocessing}
\label{sec:dataset_processing}

\paragraph{High-Level Generalist Training Data} We obtain training data for the high-level navigation module from diverse robotic navigation datasets. Since different robots may not share the same low-level action space, we align predictions across these datasets using pixel-point prediction as a unifying interface. For all data sources, we label trajectories in hindsight using camera poses at a horizon $H$ into the future. Importantly, we use poses of the robot \textit{on the ground} for all training data; this lets us specify goals in image space behind occluded points. We use known or estimated intrinsic and extrinsic matrices to project the 3D poses recorded in the datasets into 2D image trajectories.

We curate a diverse mix of datasets for navigation that spans different robot embodiments, camera perspectives, timing and weather conditions, and, importantly, different navigation capabilities and affordances. We perform several data pre-processing operations on our data that are crucial for improving model performance to the point of deployability, i.e., combining both 
short- and long-horizon trajectories, filtering data based on curvature, and empirically determining the right data mix.

\paragraph{Steerability Recipe} The textual interface of our generalist VLM lets us provide preferences expressed as text-based instructions to steer the model's predictions at test time. To train a steerable model, we augment 10\% of the data with state-of-the-art VLM annotations and co-train with two text-only visual question datasets. First, we generate 4 temporally correlated noisy versions of the ground-truth 2D trajectory $\tau$ plus a mirrored version of $\tau$. Then, we overlay all paths onto the image $I$ and use chain-of-thought prompting to ask \texttt{GPT-5-mini} to (1) describe the obstacles and terrain in the scene, (2) describe the paths, and (3) rank them based on their quality and diversity. We take the top three 2D paths and their respective descriptions, and we add them to our dataset. Finally, we co-train with data from the COCO-QA \cite{CocoQA} and Localized Narratives \cite{LocalizedNarratives} datasets to prevent forgetting.

\begin{figure*}[t]
    \centering
    \begin{subfigure}[b]{0.158\textwidth}
        \centering
        \includegraphics[width=\linewidth]{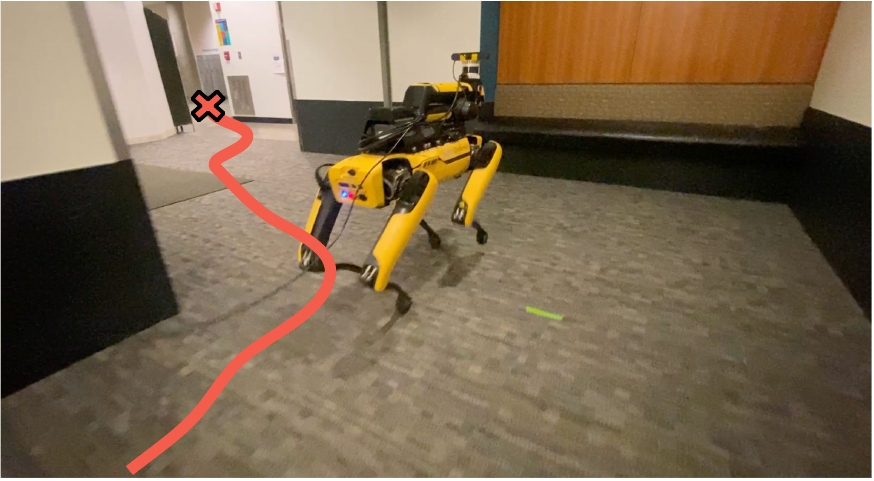}
        \caption{Hallway}
        \label{fig:hallway}
    \end{subfigure}
    \hfill
    \begin{subfigure}[b]{0.158\textwidth}
        \centering
        \includegraphics[width=\linewidth]{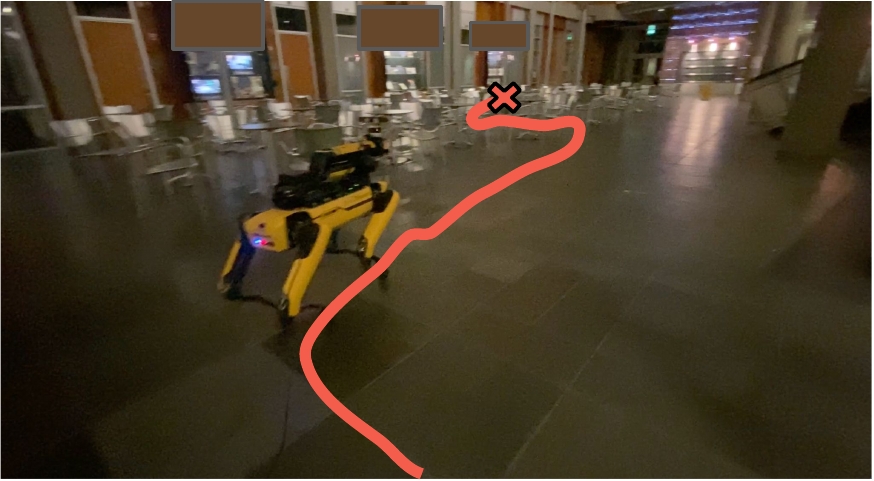}
        \caption{Atrium}
        \label{fig:atrium}
    \end{subfigure}
    \hfill
    \begin{subfigure}[b]{0.158\textwidth}
        \centering
        \includegraphics[width=\linewidth]{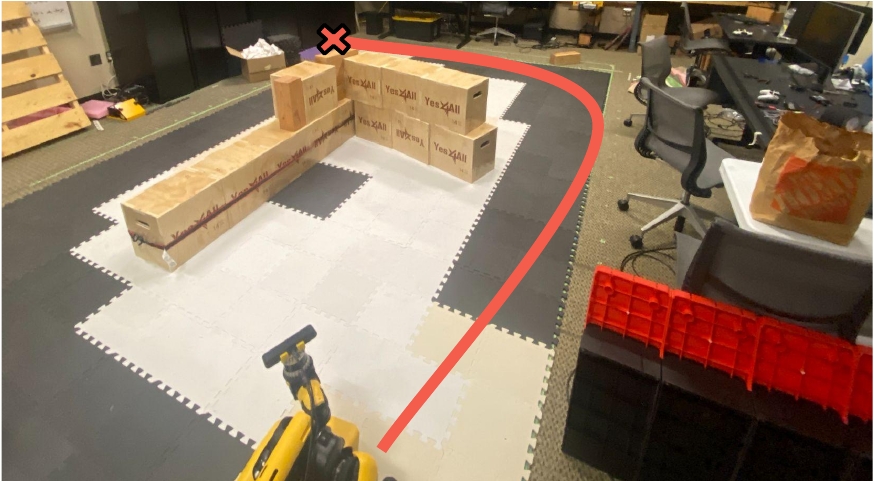}
        \caption{Lab}
        \label{fig:lab}
    \end{subfigure}
    \hfill
    \begin{subfigure}[b]{0.158\textwidth}
        \centering
        \includegraphics[width=\linewidth]{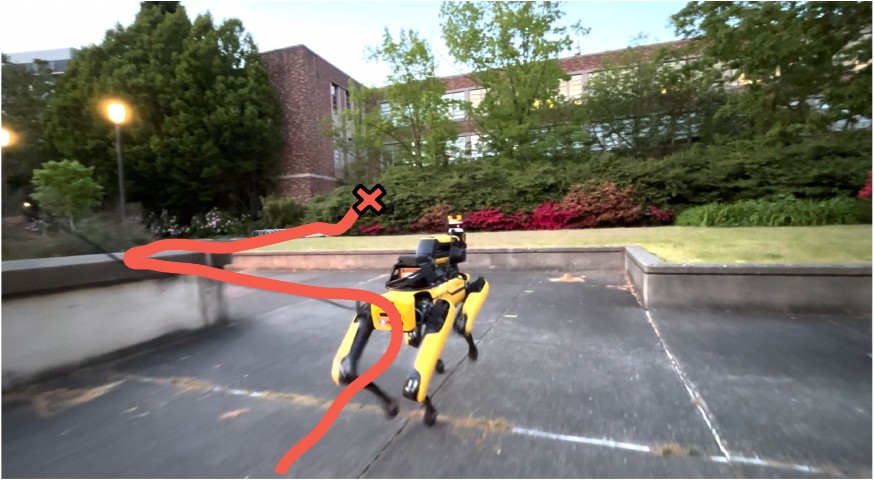}
        \caption{Campus}
        \label{fig:campus}
    \end{subfigure}
    \hfill
    \begin{subfigure}[b]{0.158\textwidth}
        \centering
        \includegraphics[width=\linewidth]{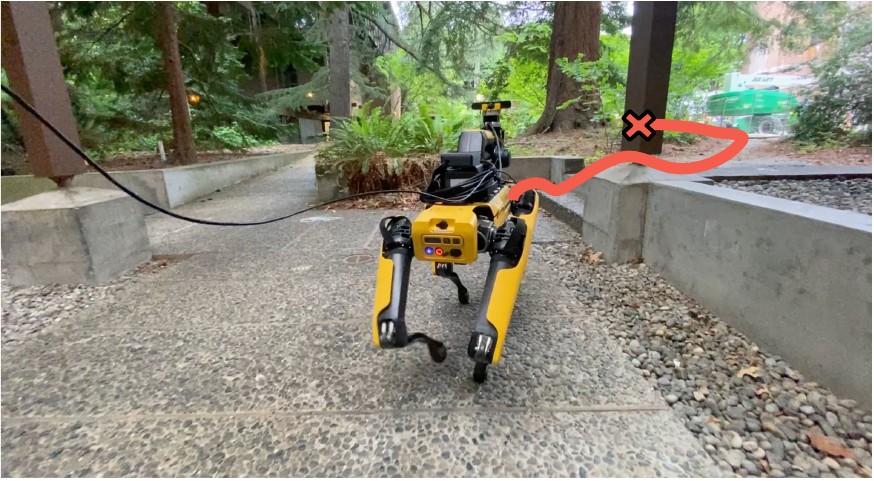}
        \caption{Forest}
        \label{fig:forest}
    \end{subfigure}
    \hfill
    \begin{subfigure}[b]{0.158\textwidth}
        \centering
        \includegraphics[width=\linewidth]{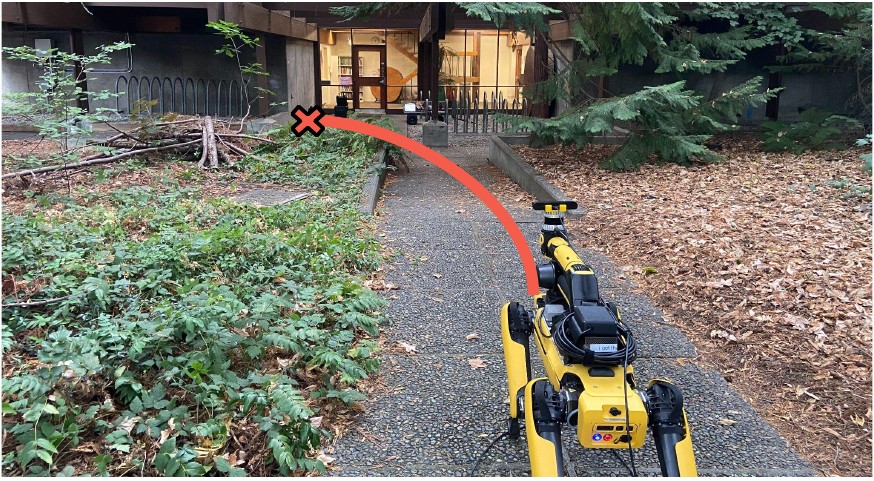}
        \caption{Down Ramp}
        \label{fig:down_ramp}
    \end{subfigure}

    \caption{We run real-world navigation experiments indoors and outdoors in unseen scenes with challenging terrain, lighting, and vegetation. Our results show that \methodname~outperforms state-of-the-art navigation foundation models and model-based baselines. }
    \label{fig:navigation_courses}
\end{figure*}

\begin{table*}[!h]
\centering
\begin{tabular}{l ccc ccc ccc ccc ccc ccc | c}
\toprule
& \multicolumn{9}{c}{\textbf{Indoor}} & \multicolumn{9}{c}{\textbf{Outdoor}} & \\
\cmidrule(lr){2-10} \cmidrule(lr){11-19}
& \multicolumn{3}{c}{Hallway} & \multicolumn{3}{c}{Atrium} & \multicolumn{3}{c}{Lab} & \multicolumn{3}{c}{Campus} & \multicolumn{3}{c}{Forest} & \multicolumn{3}{c}{Down Ramp} & \textbf{Avg. SR}\\
\cmidrule(lr){2-4} \cmidrule(lr){5-7} \cmidrule(lr){8-10}
\cmidrule(lr){11-13} \cmidrule(lr){14-16} \cmidrule(lr){17-19}
\textbf{Method} & SR & NI & T & SR & NI & T & SR & NI & T & SR & NI & T & SR & NI & T & SR & NI & T & \\
\midrule
Modular Stack & \textbf{100} & 0 & 0 & \textbf{100} & 0 & 0 & \textbf{100} & 0.2 & 0 & 0 & -- & 2 & 0 & -- & 0 & 20 & 1 & 0 & 53\\
ViPlanner     & \textbf{100} & 0 & 0 & \textbf{100} & 0 & 0 & 0 & -- & 0 & \textbf{100} & 0 & 0 & \textbf{100} & 0 & 0 & 0 & -- & 0 & 67\\
\midrule
NoMaD         & 60 & 1.3 & 1 & 0 & -- & 3 & 40 & 2 & 0 & 0 & 0 & 5 & 0 & -- & 2 & 60 & 0.7 & 0 & 27\\
NaVILA        & 20 & -- & 1 & 0 & -- & 1 & 40 & -- & 0 & 0 & -- & 0 & 0 & -- & 1 & 0 & -- & 5 & 10\\
\rowcolor{gray!15}\textbf{\methodname\ (Ours)} & \textbf{100} & 0.2 & 0 & 80 & 0.25 & 1 & \textbf{100} & 0 & 0 & 80 & 0 & 0 & \textbf{100} & 0.4 & 0 & \textbf{80} & 0.25 & 0 & \textbf{90}\\
\bottomrule
\\[-4pt]
\multicolumn{20}{l}{\small SR: Success Rate over 5 trials (\%) $\uparrow$, NI: Avg. number of interventions on successful runs [0-2] $\downarrow$, T: 3 min. timeouts [0-5] $\downarrow$}
\end{tabular}
\caption{\methodname~outperforms model-based (above the horizontal line) and end-to-end generalist navigation baselines (below the line) across a wide variety of conditions. Success Rate (SR) is computed over 5 trials for each robot-environment pair. Notably, we show that prior navigation generalists struggle to match the performance of a traditional modular stack, while \methodname~outperforms all baselines.}
\label{tab:navigation-comparison}
\vspace{-0.3cm}
\end{table*}

\subsection{Affordance Conditional Modulation}
\label{sec:valuemodulation}

\textbf{Formulation.} The high-level VLM predictions are modulated by a low-level, capability-aware affordance function, which ensures that only achievable behavior is executed on hardware. The high-level navigation policy generates a set of candidate trajectories that the robot can follow to reach the goal. To pick the trajectory candidate best suited to the specific low-level locomotion policy running on the robot, we predict an affordance score $F_\pi : M \times X \times Y \times A \to [0, 1]$ that jointly maps from an elevation map $M: \{1,2,\ldots,W\}\times\{1,2,\ldots,H\} \to \mathbb{R}$, normalized query point $x,y \in [0, 1]$ position in Euclidean space around the robot, and heading angle $a\in\{0\degree,45\degree,\ldots,315\degree\}$ to the probability that the policy $\pi$ can actually traverse $(x, y)$ in the map $M$ when heading in direction $a$. This setup is inspired by the traversability estimation literature, both in simulation \cite{frey2022locomotion, roth2025learned} and from real-world data \cite{castro2023does, mattamala2024wild}. An affordance score of $1$ indicates that the point is fully traversable, while $0$ indicates that the point is not traversable. 

This affordance function $F_\pi$ is learned via supervised learning fully in simulation by rolling out the embodiment-specific locomotion policy across a diversity of terrains. $F_{\pi}$ enables test-time modulation of predictions from the VLM and is of benefit in two situations. First, it helps to find the candidate trajectory predicted by the VLM that is best aligned with the actual capabilities of the robot. Second, it assists with filtering out potentially noisy or infeasible predictions from the VLM, e.g., if it incorrectly predicts a path through an obstacle.

\textbf{Training.} Training data for learning affordance function $F_{\pi}$ is made available by executing trajectories in \emph{simulation} over a large variety of procedurally generated terrains using the chosen low-level policy. To collect each data point, a random elevation map $M$ is spawned; following this, the agent is reset to a particular position $(x, y)$ in the simulator, the policy is executed over a short horizon in a particular direction $a$, and binary traversal success (or failure) of the low-level policy is noted. This results in a set of data points $\mathcal{D} = \{M^{(n)}, x^{(n)}, y^{(n)}, a^{(n)}, s^{(n)}\}_{n=1}^N$, where $M^{(n)}\in \mathbb{R}^{W\times H}$ is a local elevation map, $(x^{(n)}, y^{(n)})$ is the queried agent position, $a^{(n)}\in\{0\degree,45\degree,\ldots,315\degree\}$ is the heading direction, and $s^{(n)}\in\{0,1\}$ is a label representing success or failure of the trajectory. Given this training data $\mathcal{D}$, we train an affordance function $F_{\pi}$, represented as an MLP by minimizing a standard binary cross-entropy loss $\ell$ -- $\mathcal{L} = \min_{F_\pi} \mathbb{E}_{M, x, y, a, s \sim \mathcal{D}}\left[ \ell\left(F_\pi(M, x, y, a), s\right)\right]$.

\subsection{Deployment} 
\label{sec:deployment}

The navigation missions are defined given a series of GPS waypoints or 3D coordinates in the world frame, which are converted to 2D points in the image to be passed as input to the high-level VLM. During deployment, the VLM is first queried on the current image $I$ and a text-encoded 2D goal coordinate $g_t$ to obtain a set of viable paths $p_1, p_2, \dots, p_K$ in pixel space. Each pixel-space path $p_i$ is then projected into world positions of the robot in the ground plane along each path: $\tau^w_i = \left[(x_0, y_0)^i, \dots, (x_H, y_H)^i\right]_{i=1}^K$ to query affordances. The affordance of each candidate path is then computed using this sequence of points along with the local elevation map $M$ to query $F_\pi$, thereby obtaining a pointwise affordance score for each path: $\left[F_\pi(M, x_0, y_0, a_0)^i, \dots, F_\pi(M, x_H, y_H, a_H)^i\right]_{i=1}^K$. Finally, since a path is blocked if even one of its elements is blocked, a cumulative affordance is computed as the minimum affordance score along each path: $F^c(p^w_i) = \min \left[F_\pi(M, x_0, y_0, a_0)^i, \dots, F_\pi(M, x_H, y_H, a_H)^i\right].$ Intuitively, paths $\tau^w_i$ with higher affordances are better, while low-affordance paths are unlikely to be successfully navigated using the low-level policy $\pi$.  Given this per-path measure of cumulative affordance $F^c(p^w_i)$, we can select a single trajectory to execute on the robot greedily by choosing the trajectory with the highest affordance, or we can sample  with soft sampling to allow for some stochasticity in path selection: $\hat{\tau}^w \sim \text{Softmax}\left( \frac{F(\tau^w_1)}{\beta}, \frac{F(\tau^w_2)}{\beta}, \ldots, \frac{F(\tau^w_k)}{\beta} \right)$.

This modulation results in a sample path $\hat{\tau}^w$ that can then be executed on the robotic hardware by commanding waypoints to the low-level policy. During deployment, we assume access to a low-level, velocity- or position-conditioned locomotion controller for our real-world platforms. We use the predictions of the high-level VLM in a receding horizon control fashion, where it predicts $k=5$ waypoints but uses only the first $m$ waypoints predicted by the high-level controller before replanning, where $m<k$ is a tunable parameter. If the goal coordinate is not in the image frame, the robot rotates in place until the goal is back in the image before replanning.

\section{Experiment Results}
\label{sec:result}

Out experiments evaluate the following research questions. (1) Is our hierarchical navigation method competitive with other navigation baselines in unseen environments? (2) Does our navigation method support cross-embodiment navigation? (3) Is \methodname~steerable? (4) Do we benefit from having a high-level generalist VLM compared to having a robot-specific navigator? (5) Do we benefit from low-level affordance modulation for single-robot navigation? We first describe the setup of our experiments and then walk through results pertaining to each question.

\subsection{Experiment Setup}

To validate the claims in this work, we test the methodology on two robotic platforms: 

\textbf{1. Legged: Boston Dynamics Spot.}
We evaluate performance on the BD Spot Robot using the built-in locomotion controller (capable of traversing ramps, stairs, and other terrains) as the low-level policy.  

\textbf{2. Wheeled: UW Hound Robot.} To test transfer across embodiments, we also consider a second robot, the UW Hound~\cite{talia2023hound}. Importantly, the Hound uses the same high-level VLM planner, but we simply vary the low-level affordance function and controller. 

\textbf{Simulation Environment.} We build our simulation environment to learn the affordance function on Isaac Lab. We use a perceptive RL policy trained with reinforcement learning in simulation \cite{mittal2023orbit} as a proxy for the built-in BD Spot policy. To learn perceptive affordance functions that transfer well to real world, we must provide a wide diversity of terrains in simulation; during real-world deployment, there are often more distractors in the environment, such as furniture or vegetation, that must be modeled for proper sim-to-real transfer. To add diversity to our simulation environments, we generated inter-connected structures with stairs and ramps using wave function collapse. Additionally, to model irregular patterns, we used cellular automata to generate smooth, uneven terrains.

\subsection{Is \methodname~ a capable navigation system in the real world?}
\label{sec:real_world_nav}
\begin{figure}[!t]
    \centering
    \includegraphics[width=0.8\linewidth]{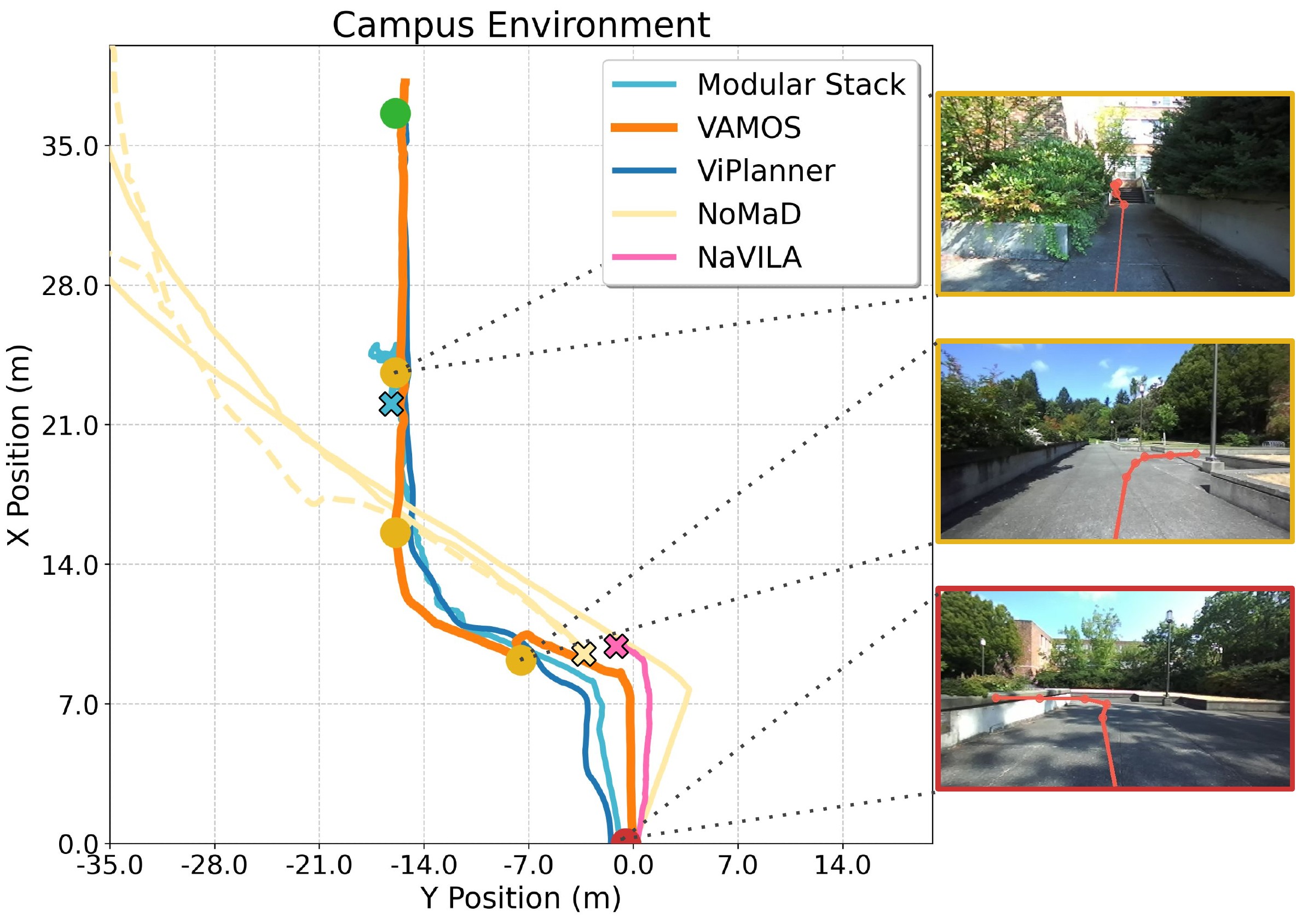}
    \caption{Top-down map showing paths taken by different methods from start (red) to goal (green) through waypoints (yellow). \methodname~achieves long-horizon, precise navigation. Right: predicted and selected paths when replanning after reaching a waypoint. Dotted lines show returns to the last completed waypoint after interventions; X's mark baseline failures or timeouts.}
    \label{fig:outdoornav}
    \vspace{-0.7cm}
\end{figure}

We compare performance between our method and other state-of-the-art baselines in terms of navigation capabilities in real-world, unseen, indoor and outdoor environments (Fig. \ref{fig:navigation_courses}). The chosen baselines are (1) a geometric model-based modular navigation stack similar to \cite{meng2023terrainnet}, (2) ViPlanner~\cite{roth2024viplanner}, a learned geometric and semantic planner, (3) NoMaD~\cite{sridhar2024nomad}, a navigation foundation model, and (4) NaVILA \cite{cheng2024navila}, a navigation VLA. We focus on a short- to medium-horizon range for goal navigation, where the goal position is specified in 3D global coordinates. To reach long-range goals, we generate waypoints to the goal every $\sim10$ meters (Fig. \ref{fig:outdoornav}).

The ``Hallway" course ($\sim20m$) tests the ability to navigate down narrow corridors with tight turns. The ``Atrium" course ($\sim20m$) measures the ability to navigate cluttered open scenes in low light. The ``Lab" ($\sim5m$) course tests the ability to navigate to a point occluded by a large irregular obstacle. The ``Campus" ($\sim40m$) course tests the ability to navigate long distances, including going up a 7-step staircase. The ``Forest" ($\sim20m$) course tests the ability to navigate in vegetated environments that including stairs; rooted and vegetation-covered terrain; irregular concrete paths; and paths with overhanging vegetation. Finally, the ``Down Ramp" ($\sim15m$) course tests the ability to navigate to a point below the start pose, evading foot-snaring vines.

We present the results in Table~\ref{tab:navigation-comparison}. \textit{\methodname~achieves higher average success rate across all courses, performing well across all conditions, which no other baseline does. }

In indoor environments, \methodname~performs on par with the modular stack and ViPlanner, with the exception being the more challenging "Lab" course, where it outperforms all baselines. This is because the inferred geometric cost-maps indoors are clean and easy to plan against. However, two generalist baselines, NoMaD and NaVILA, struggle to generalize out-of-distribution, even though they were both trained using indoor data similar to our data mix, and mainly navigate in straight lines or bounce off walls. We credit \methodname's superior performance to our usage of 2D trajectories, which have been shown to maintain more of the pre-trained VLM's generalization capabilities~\cite{li2025hamster}. 

\methodname~also excels in outdoor urban and off-road environments. Neither the modular stack nor the generalist baselines perform well outdoors. The geometric modular stack fails at the interface of perception and planning, where inaccurate perception leads to downstream failures. The generalist baselines fail because in more open environments, they mainly walk in straight lines. ViPlanner performs well due to its well-tuned geometric and semantic perception integration. However, in both the ``Lab" and ``Down Ramp" environments, which are challenging due to large geometric obstacles that require long-term planning, ViPlanner fails to reason about long-term outcomes. \textit{These experiments highlight \methodname's rich geometric and semantic reasoning capabilities, resulting in a significantly higher overall average  success rate (90\%) compared to the baselines.}

\subsection{Does \methodname~support cross-embodiment navigation?}
\label{sec:cross-embodiment}

\begin{figure}[t]
\centering
\includegraphics[width=0.8\linewidth]{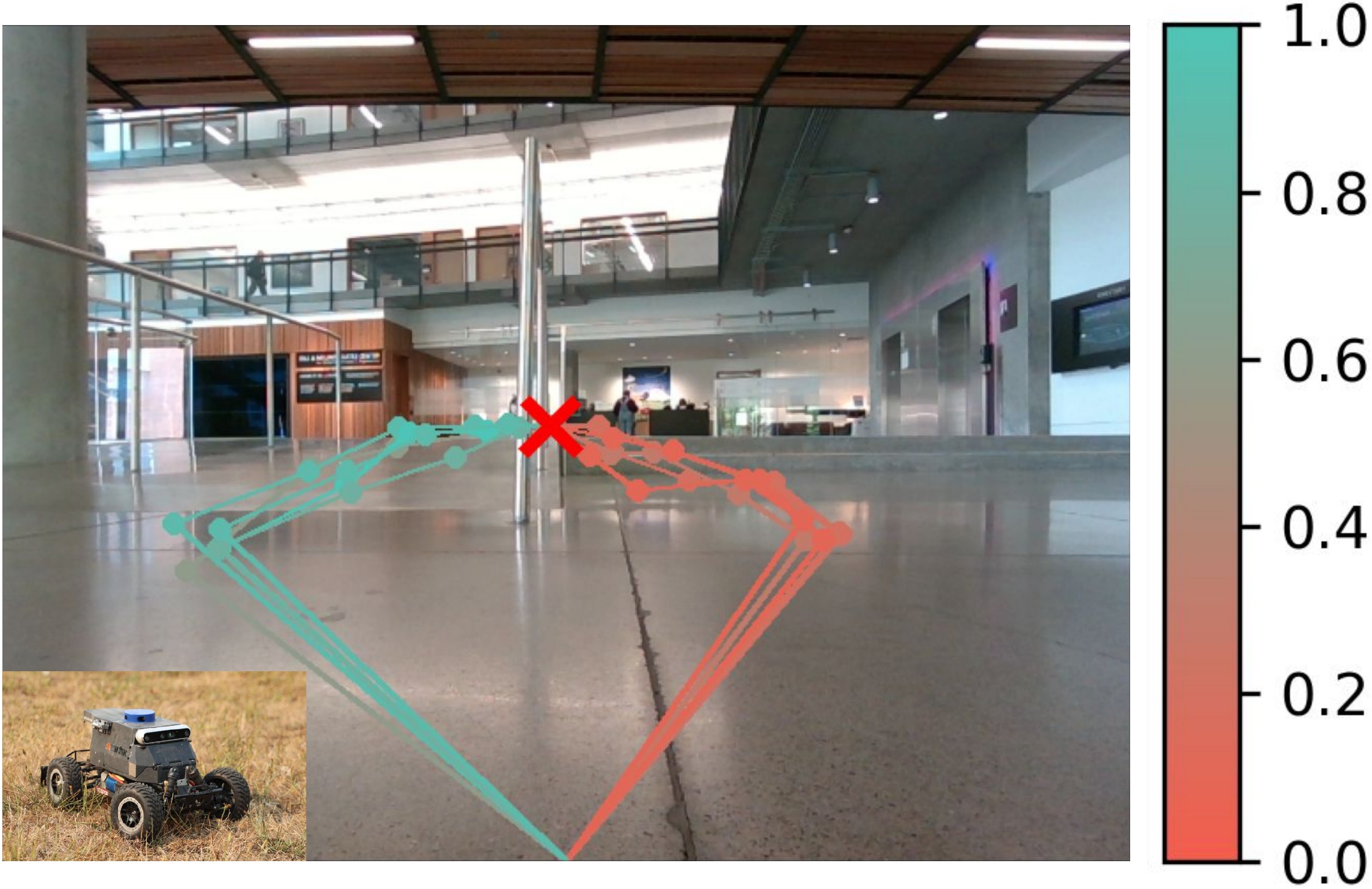}
\caption{We evaluate the cross-embodiment capabilities of \methodname~on a wheeled robot, Hound, where the goal (red X) is to reach an elevated floor through either a ramp or stairs. We show 10 candidates predicted by the VLM and their corresponding affordance score. Re-ranking with the affordance function enables higher success rates in cross-embodied navigation as shown in Table \ref{tab:robot-navigation-choices}.}
\label{fig:traversability-choice}
\end{figure}

\begin{table}[t]
\centering
\setlength{\tabcolsep}{4pt}
\renewcommand{\arraystretch}{1.15}
\begin{tabularx}{1.0\linewidth}{@{}l *{2}{>{\centering\arraybackslash}X} c *{2}{>{\centering\arraybackslash}X} c@{}}
  \toprule
  & \multicolumn{3}{c}{\textbf{Spot}} & \multicolumn{3}{c}{\textbf{Hound}} \\
  \cmidrule(lr){2-4} \cmidrule(lr){5-7}
   & Stairs & Ramps & SR & Stairs & Ramps & SR \\
  \midrule
  No Modulation   & 4/10 & 6/10 & 100\% & \textcolor{vamos_red}{4/10} & \textcolor{vamos_blue}{6/10} & 60\% \\
  With Modulation & 8/10 & 2/10 & 100\% & \textcolor{vamos_red}{1/10} & \textcolor{vamos_blue}{9/10} & \textbf{90\%} \\
  \bottomrule
\end{tabularx}
\caption{Affordance modulation maintains Spot’s perfect performance and improves Hound’s overall success rate from 60\% to 90\%, enabling cross-embodied navigation. Counts show per-terrain path choices (teal = success, red = failure) over 10 runs.}
\label{tab:robot-navigation-choices}
\end{table}

\begin{table}[h!]
\centering
\setlength{\tabcolsep}{4pt}
\renewcommand{\arraystretch}{1.15}
\begin{tabularx}{0.6\linewidth}{@{}l >{\centering\arraybackslash}X >{\centering\arraybackslash}X@{}}
  \toprule
  \textbf{Method} & \textbf{Spot} & \textbf{Hound} \\
  \midrule
  ViPlanner & 100\% & 0\% \\
  \methodname & 100\% & \textbf{90\%} \\
  \bottomrule
\end{tabularx}
\caption{\methodname~outperforms the best navigation baseline in cross-embodiment tasks, selecting ramps vs.\ stairs aligned with robot capabilities via its affordance model (N=10).}
\label{tab:cross-embodiment}
\vspace{-0.7cm}
\end{table}

\begin{figure}
    \centering
    \includegraphics[width=1.0\linewidth]{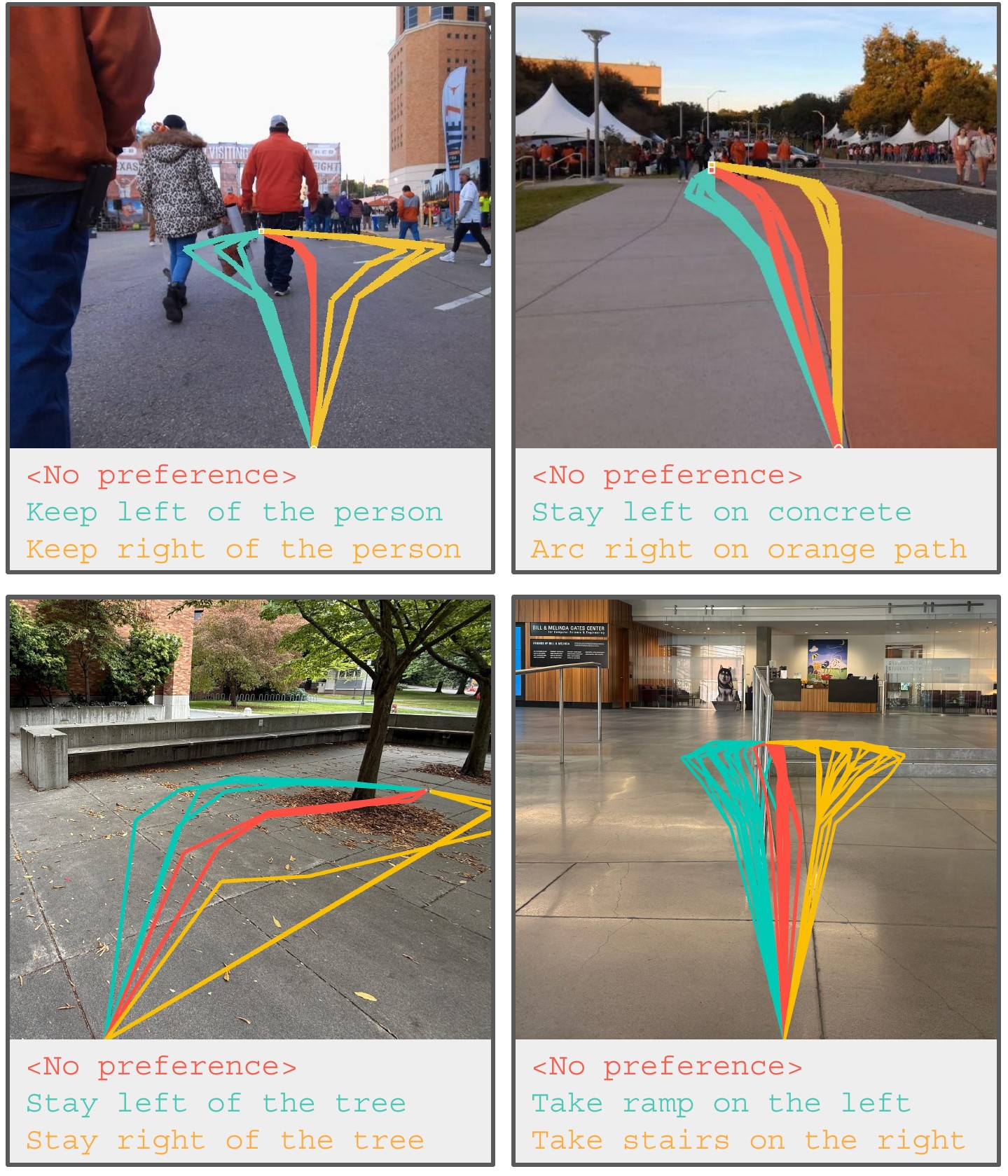}
    \caption{\methodname~is steerable through natural language preferences appended to its goal coordinate specification. Different preferences are indicated by the shown natural language prompts and depicted using different colors.}
    \label{fig:steerability}
    \vspace{-3mm}
\end{figure}

We evaluate the cross-embodiment capabilities of our method on a simple test environment consisting of a staircase and a ramp, side-by-side, leading to an elevated floor, as shown in Figure \ref{fig:traversability-choice}. We use the same high-level planner for both Spot and Hound robots, and we swap only the embodiment-specific affordance module. First, we show that affordance modulation lets the same VLM predictor be used effectively with two different robot embodiments, enabling navigation for both platforms. As we show in Table~\ref{tab:robot-navigation-choices}, the same VLM \emph{with} affordance modulation enables accurate navigation for both legged and wheeled platforms, taking specific robot capabilities into account. In this case, the wheeled robot can only take the ramp, while the legged robot can succeed on both stairs and ramps. In contrast, executing VLM predictions without affordance modulation often results in predictions that are not achievable under the current low-level embodiment. \footnote{To improve multimodal generation in this experiment, we collected 50 static images with slight pose variations from each robot in that environment, labeled each with a path going up stairs and a path going up ramps, and then generated 10 noisy samples per hand-drawn trajectory to generate the dataset that we used to finetune the base \methodname VLM planner. This helped more clearly illustrate the differentiation provided by the affordance function.}

Compared to the best performing method in Table \ref{tab:navigation-comparison}, ViPlanner, we show that our method achieves almost perfect success rates on both embodiments, while ViPlanner fails when deployed on Hound, as shown in Table \ref{tab:cross-embodiment}. By swapping affordance models that are cheap to train and run, we obtain performant cross-embodiment navigation.

\subsection{Is \methodname~steerable via natural language?}
\label{ref:steerable}
We evaluate the steerability of our model qualitatively and quantitatively. In Figure \ref{fig:steerability}, we show examples of the 2D paths predicted by \methodname~with and without preferences appended to the text input that  encodes the goal coordinate. As shown in Figure \ref{fig:steerability}, we can adapt the output trajectories to follow a particular direction (left or right) or to take a particular terrain (stairs, ramps, or grass planters). Using VLM-as-a-judge (ChatGPT 5) on the bottom-right image in Figure \ref{fig:steerability} , we obtain 20/20 preference alignment when specifying which path to take for both the ramps and the stairs compared to the original trajectories without pre-specified preferences.

\subsection{Does the high-level VLM generalist provide benefits over a robot-specific navigator?}
\label{sec:pooling}

\begin{figure}[!h]
    \centering
    \includegraphics[width=0.85\linewidth]{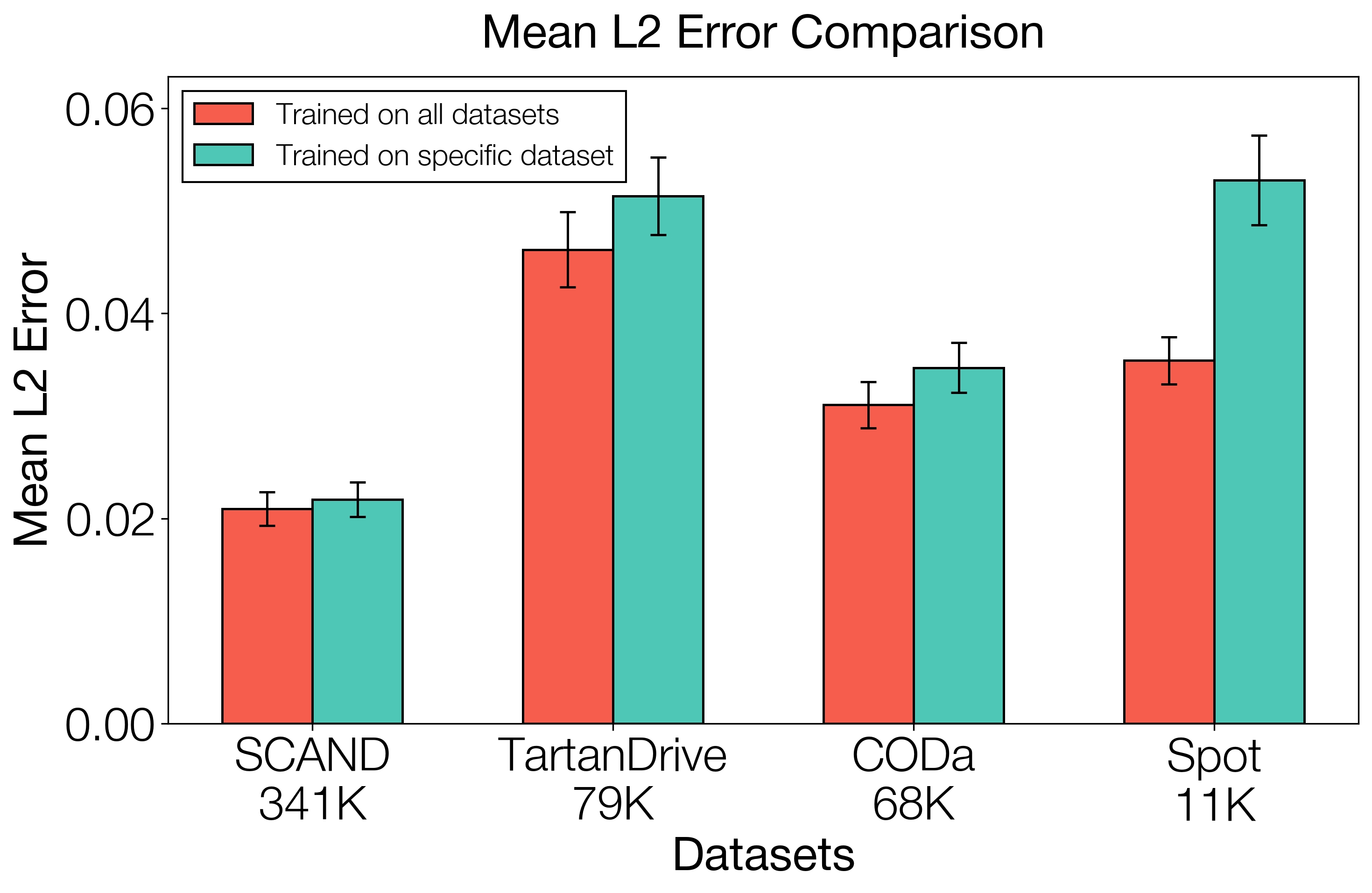}
    \caption{Pooling data across all robot datasets (red) improves model performance compared to training specialist navigation models on individual robot datasets (teal). We evaluate over the entire validation set. Error bars represent 95\% CI.}
    \label{fig:poolingviz}
    \vspace{-0.4cm}
\end{figure}

To understand whether training a generalist VLM policy is actually beneficial, we perform an analysis of offline model performance. Specifically, we aim to answer whether pooling data from the heterogeneous datasets in Figure~\ref{fig:dataset_depiction} is beneficial compared to simply training the model on single, robot-specific datasets. We compare the performance of the high-level VLM predictor on path prediction across mean L2 prediction error as a metric. Specifically,  we compare the performance of a model trained on a pooled dataset across all the datasets mentioned in Figure \ref{fig:dataset_depiction} to the performance of a model trained on each individual dataset. The results in Figure~\ref{fig:poolingviz} indicate that pooling data results in better performance than training on specific datasets. 

\begin{figure}[t]
\centering
\includegraphics[width=0.8\linewidth]{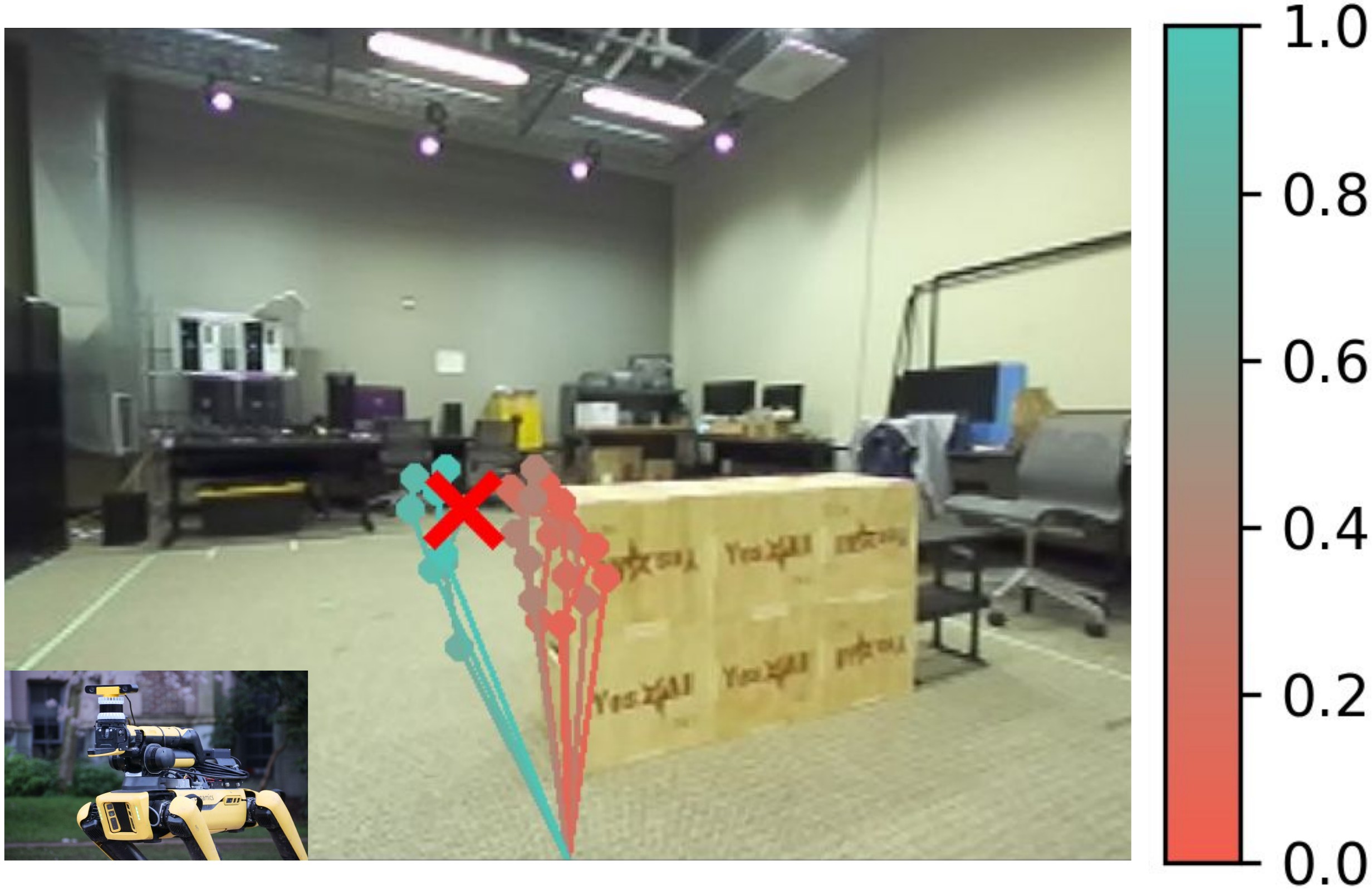}
\caption{The affordance function also helps with filtering out noisy VLM predictions in single-embodiment OOD settings. In this example, it filters out the paths predicted by the VLM that go through obstacles to reach the goal (red X), leading to higher success rate (Table \ref{tab:value-function-comparison}). }
\label{fig:traversability-image}
\end{figure}

\begin{table}[h!]
\centering
\setlength{\tabcolsep}{4pt}
\renewcommand{\arraystretch}{1.15}
\begin{tabularx}{0.6\linewidth}{@{}l >{\centering\arraybackslash}X@{}}
  \toprule
  \textbf{Condition} & \textbf{Success Rate} \\
  \midrule
  No Modulation & 20.0\% \\
  With Modulation    & \textbf{60.0\%} \\
  \bottomrule
\end{tabularx}
\caption{Affordance modulation reduces high-level VLM prediction errors in cases where the VLM predicts paths that violate the robot's capabilities or physics. Success rate is over 10 runs.}
\label{tab:value-function-comparison}
\end{table}

\subsection{Do we benefit from low-level affordance modulation for single-robot navigation?}
\label{sec:modulation}

Next, we evaluate whether modulation with the affordance function can improve model performance with a single embodiment by correcting for VLM errors. We show quantitatively in Table~\ref{tab:value-function-comparison} that the VLM performance without modulation can make mistakes in OOD settings, such as going through obstacles, that are corrected by the affordance function modulation. The same can be seen qualitatively in Figure~\ref{fig:traversability-image}, where affordance modulation prevents the execution of catastrophic paths suggested by the VLM.

\begin{figure}[t]
    \centering
    \begin{subfigure}[b]{0.31\columnwidth}
        \centering
        \includegraphics[width=\linewidth]{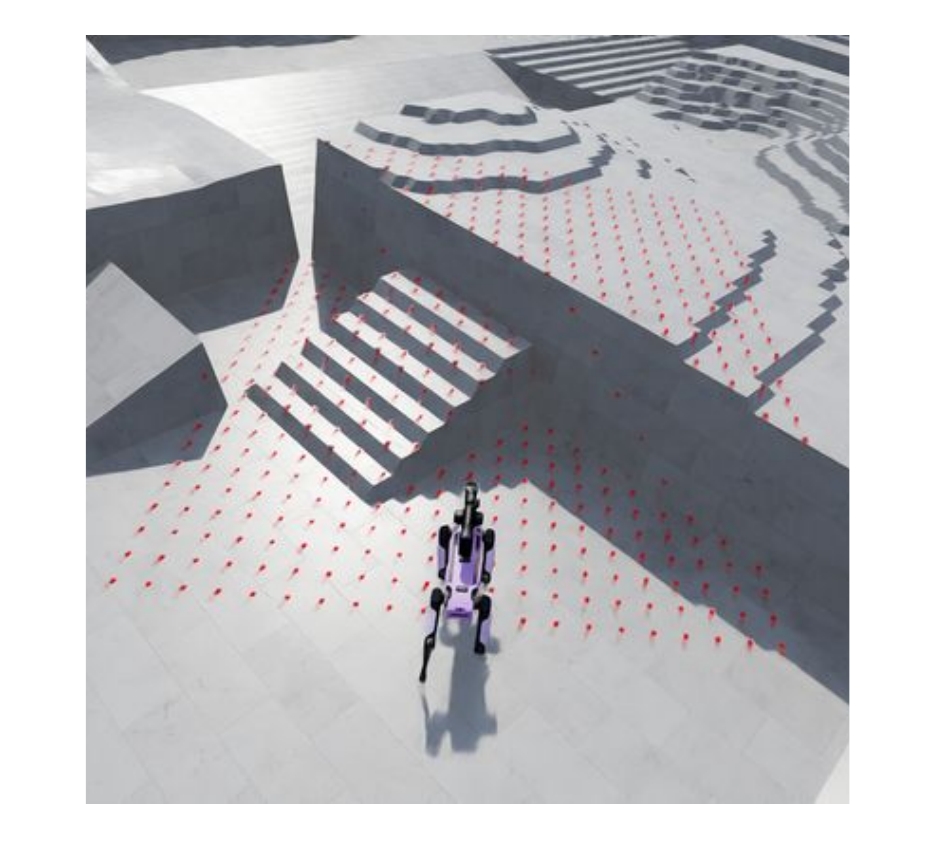}
        \caption{Scene Geometry}
        \label{fig:scene_geometry}
    \end{subfigure}
    \hfill
    \begin{subfigure}[b]{0.31\columnwidth}
        \centering
        \includegraphics[width=\linewidth]{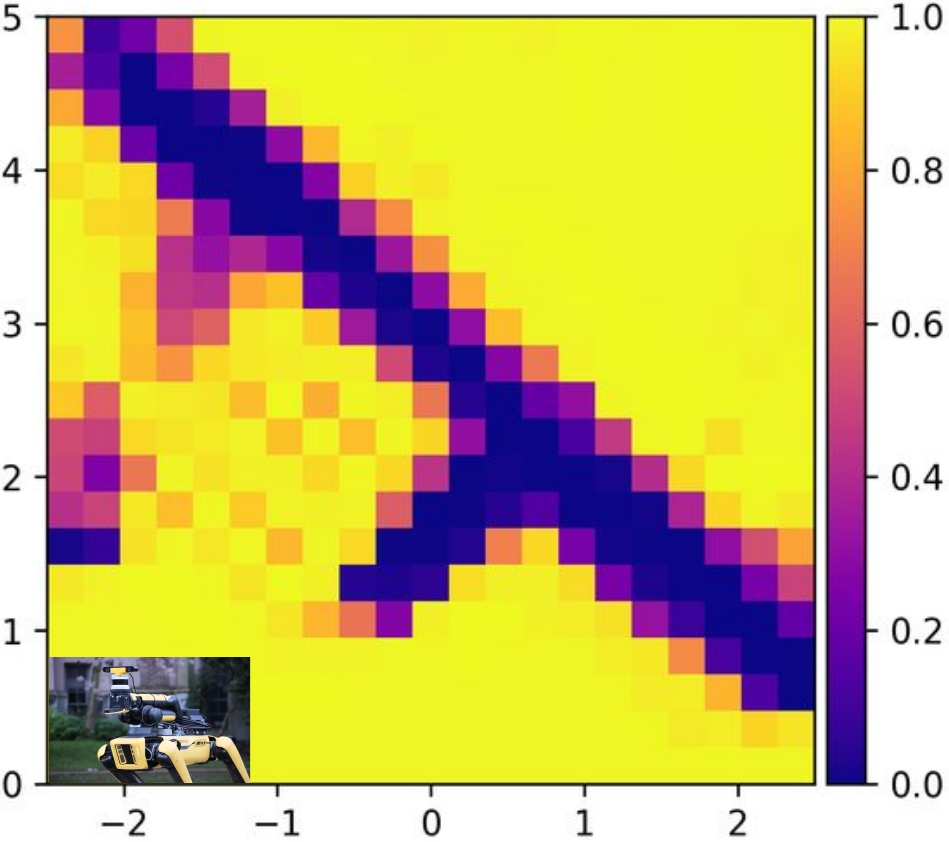}
        \caption{Spot Affordance}
        \label{fig:spot_affordance}
    \end{subfigure}
    \hfill
    \begin{subfigure}[b]{0.31\columnwidth}
        \centering
        \includegraphics[width=\linewidth]{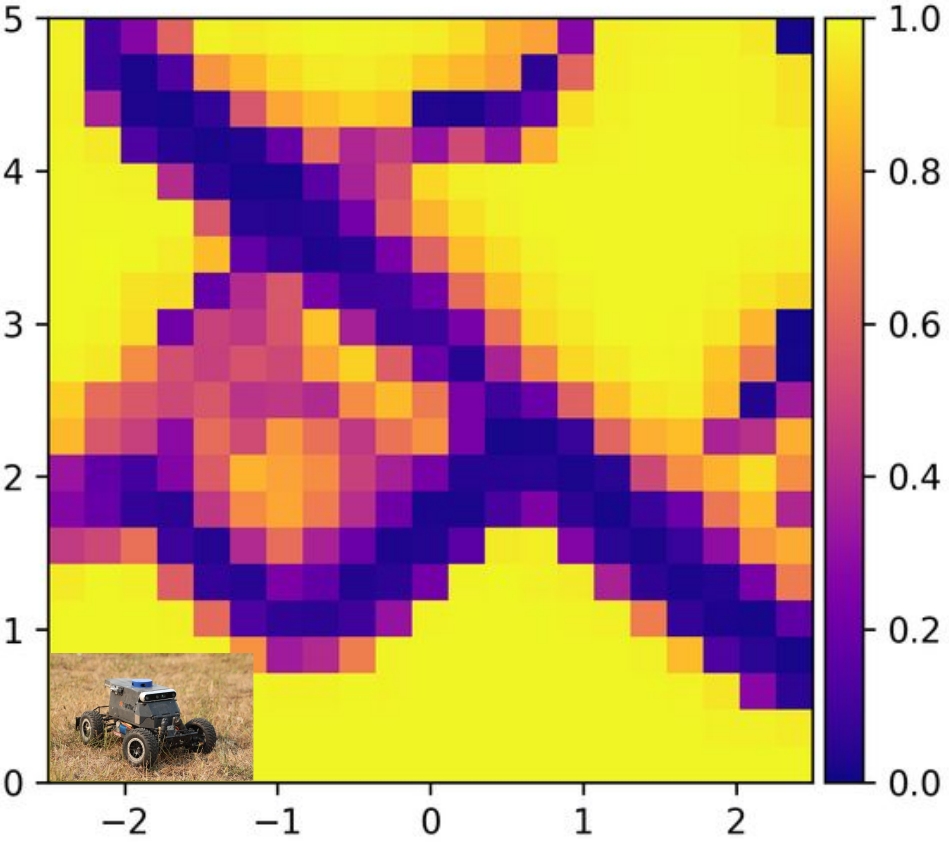}
        \caption{Hound Affordance}
        \label{fig:hound_affordance}
    \end{subfigure}

    \caption{The affordance function indicates that the Spot robot can ascend stairs, but the wheeled Hound cannot (yellow signifies high-affordance score). Both robots cannot traverse tall obstacles (e.g., the wall has a low-affordance score).}
    \label{fig:traversability}
\end{figure}

Finally, we visualize the affordance function in Figure~\ref{fig:traversability}. We see that it naturally captures the geometry of the environment and the particular agent's capabilities. Projecting this affordance function onto the VLM predictions prevents mistakes like navigating directly into obstacles.

\section{Conclusion}
\label{sec:conclusion}
We presented \methodname, a technique for general-purpose navigation using vision-language models. The central idea in this work is to combine diverse, heterogeneous datasets for training a hierarchical VLA model. The high-level VLM planner predicts candidate navigation paths as 2D pixel paths. This output is modulated by a low-level affordance model that enables capability- and embodiment-aware navigation on deployment. We show significantly improved performance over both model- and learning-based baselines in our extensive real-world navigation experiments. The resulting methodology provides a step towards open-world, general-purpose navigation agents that can reason both geometrically and semantically about how to act in the world.

\section*{Acknowledgements}
This research was partly funded by the Army Research Lab (ARL) and compute provided by the University of Washington Hyak program. The authors would like to thank Xiangyun Meng and Yi Li for early discussions and feedback. The authors would also like to thank Khimya Khetarpal, Daphne Chen, Brady Moon, Gokul Swamy, Alex Stephens, and Swapnil Pande for presentation feedback, as well as other members of Robot Learning Lab and WEIRD Lab at University of Washington for feedback and support. Finally, the authors would like to thank the authors of the baselines we compared against for providing their code.

\bibliographystyle{unsrt}
\bibliography{references}

\clearpage
\newpage
\appendices

\section{High-Level Training Details}
\subsection{Hyperparameters and Compute}

We present all our training hyperparameters for the high-level VLM in Table \ref{tab:hyperparameters_concise}. We find that training for multiple epochs lends to rapid overfitting, so we train our model for 1 epoch using an Nvidia L40 node of 8 GPUs with a per-device batch size of 8 for about 5 hours. Notably, we find that it is possible to fine-tune our model with LoRA on a consumer-grade Nvidia RTX 4090 GPU, albeit with a much smaller per-GPU batch size of 2. We take advantage of state-of-the-art training infrastructure for large language models (LLMs) by integrating our training with the HuggingFace ecosystem, using the TRL library \cite{vonwerra2022trl} with data-parallelism implemented by the \texttt{accelerate} library \cite{accelerate}.

\begin{table}[htbp]
\centering
\caption{Key Training Hyperparameters}
\label{tab:hyperparameters_concise}
\begin{tabular}{@{}ll@{}}
\toprule
Hyperparameter                 & Value                                                                    \\
\midrule
Base Model                     & \texttt{google/paligemma2-3b-pt-224}                                     \\
Seed                           & 42                                                                       \\
Optimizer                      & \texttt{adamw}                                                    \\
Learning Rate                  & 1e-4                                                                     \\
Adam $\beta_1$                 & 0.9                                                                      \\
Adam $\beta_2$                 & 0.999                                                                    \\
Adam $\epsilon$                & 1e-8                                                                     \\
Weight Decay                   & 1e-5                                                                     \\
Max Grad Norm                  & 1.0                                                                      \\
LR Scheduler                   & Cosine                                                                   \\
Warmup Ratio                   & 0.1                                                                      \\
Num Train Epochs               & 1                                                                        \\
Batch Size (per device)        & 8                                                                        \\
Gradient Accumulation Steps    & 1                                                                        \\
Num GPUs    &                     8                                                    \\
Effective Batch Size           & 64                                           \\
Precision                      & \texttt{bfloat16}                                                        \\
Max Sequence Length            & 2048                                                                     \\
Data Packing                   & True                                                                     \\
\midrule
\multicolumn{2}{c}{LoRA Specific Parameters (PEFT)} \\
\midrule
LoRA R (Rank)                  & 16                                                                       \\
LoRA Alpha                     & 16                                                                       \\
LoRA Dropout                   & 0.05                                                                     \\
LoRA Target Modules            & \begin{tabular}[t]{@{}l@{}}\texttt{q\_proj, k\_proj, v\_proj, o\_proj,} \\ \texttt{gate\_proj, up\_proj, down\_proj}\end{tabular} \\
\bottomrule
\end{tabular}
\end{table}

\subsection{Dataset Preparation and Mixtures}

\paragraph{Dataset Processing} We perform several data pre-processing operations on our data that allows us to obtain higher-quality data for training. Notably, scaling up navigation datasets naively leads to a lot of data where the navigator mostly walks or drive straight. We balance short and long-range trajectories by sampling from two different horizons at a 50\% ratio, which increases the diversity in paths while maintaining effective short-range navigation.  Given that much of the data in these datasets is highly-correlated, we also filter the number of trajectories to maintain the most salient examples. To do this, we rank trajectories based on curvature, defined as the ratio between the ground-truth trajectory length and the straight-line distance to the goal, i.e., $c = \frac{\sum_{i=1}^{k-1} \lVert w_t^{i+1} - w_t^i \rVert}{\lVert w_t^k - w_t^1 \rVert}$, where $\tau_t = \{w_t^1, \ldots, w_t^k\}$, and we select the top $n$ data points based on curvature, where $n$ varies based on dataset. The odometry in these datasets can be noisy, so we also filter out the top 3\% of trajectories based on this curvature metric to reject noisy samples. Finally, we align the 2D image coordinate representation of the goal to the tokenization scheme of the pre-trained PaliGemma 2 model -- in particular, location tokens are represented using 1024 discrete location tokens (\texttt{<loc0000>} to \texttt{<loc1023>}) corresponding to binned normalized image coordinates. We convert the goal to a text instruction of the form \texttt{"Navigate to x=<locXXXX>, y=<locYYYY>"}, which then gets tokenized and passed into the model alongside the image. If a natural-language preference is specified, we append this preference to this string, e.g., \texttt{"Navigate to x=<locXXXX>, y=<locYYYY>.Stay on the right of the people."}

\paragraph{Dataset Mixtures:} In early experiments, we found training with all the data, or training with a uniformly subsampled percentage of all the data performed worse than our data mix detailed in Table \ref{tab:data_mix}. We arrived to this mix heuristically: we found that the SCAND dataset \cite{karnan2022socially} contains high-quality, diverse data, so we keep a high proportion of it, whereas the CODa dataset \cite{zhang2024towards} is very repetitive, covering very similar scenes throughout the dataset, although at higher variations of weather and lighting conditions, so we down-weight it. We keep the full datasets for the TartanDrive \cite{sivaprakasam2024tartandrive} and in-domain Spot datasets given their relatively-small size, although we filter out trajectories with noisy odometry from the TartanDrive dataset. Finally, we consider only the Spot subset of data from the SCAND dataset \cite{karnan2022socially} given that we could not obtain accurate camera parameters for the Jackal subset.

\begin{table}[t]
\centering
\begin{tabular}{@{}lrr@{}}
    \toprule
    \textbf{Dataset} & \textbf{Hours} & \textbf{Used (\%)} \\
    \midrule
    SCAND~\cite{karnan2022socially} & 19.5 & 351.2K (50\%) \\
    CODa~\cite{zhang2024towards} & 7.8 & 70.5K (25\%) \\
    TartanDrive 2~\cite{sivaprakasam2024tartandrive} & 2.2 & 79.1K (100\%) \\
    Spot & 0.3 & 11.2K (100\%) \\
    \midrule
    Human Sketch (FT) & -- & 2K (100\%) \\
    \midrule
    Total & 29.8 & 514K \\
    \bottomrule
\end{tabular}
\caption{\textbf{Dataset mix used for training high-level navigation.} All datasets include odometry annotations. Human sketch annotations enable few-shot adaptation (Section~\ref{sec:cross-embodiment}).}
\label{tab:data_mix}
\end{table}

We also experimented with two additional sources of non-robot data: videos processed with monocular tracking \cite{karaev2024cotracker} or structure-from-motion algorithms \cite{murai2024mast3r, duisterhof2024mast3r}, and data collected with paired odometry using an iPhone and its built-in odometry estimation through ARKit, similar to \cite{ha2024umi}. For unlabeled egocentric videos, we obtain estimated camera poses using the CoTracker video tracker model \cite{karaev2024cotracker}, similar to \cite{schmittle2025longrangenavigatorlrn}, which tracks a grid of 2D points on a video. To obtain trajectories using CoTracker, we run egocentric videos in reverse and track a subset of the grid of points sampled on the ground in front of the camera to obtain a sequence of traversed points. We collected a dataset of 3.7 hours, with 133K data points, of in-domain walking data collected with an Insta360 fisheye camera. We found that adding this data to our data mix hurt performance. CoTracker struggles with maintaining trajectories behind occlusions, which leads to shorter ground-truth trajectories, noisy data, and mostly straight paths. We also experimented with Mast3r-SLAM \cite{murai2024mast3r}, and while it handled occlusions better, processing long-horizon trajectories was too computationally inefficient. 

The second source of data we experimented with was odometry-labeled videos collected with an iPhone and labeled with ARKit. This allowed us to collect data at a much higher speed than through robot data collection. Most of our efforts focused on collecting data to improve the multi-modal capabilities of the robot, by starting at similar positions but taking different paths to reach the goal. We focused on collecting data going up stairs and ramps to support our experiment in Section \ref{sec:cross-embodiment}. We collected a dataset of 81K data points (about 2.3 hours). However, we found that using the entirety of this data lead to more twisty predicted trajectories throughout, and hurt quantitative metrics. We believe this data collection approach to be quite promising, both due to its ease of scalability and potential for collecting targeted data beyond what is usually found in internet and existing robotic datasets. We leave finding better ways to select data mixes from diverse sources such as the ones described in this section \cite{hejna2024re} for future work. 

\section{Qualitative Results}

\begin{figure*}[t]
    \centering

    \begin{subfigure}[b]{0.48\textwidth}
        \centering
        \includegraphics[height=5.625cm, keepaspectratio]{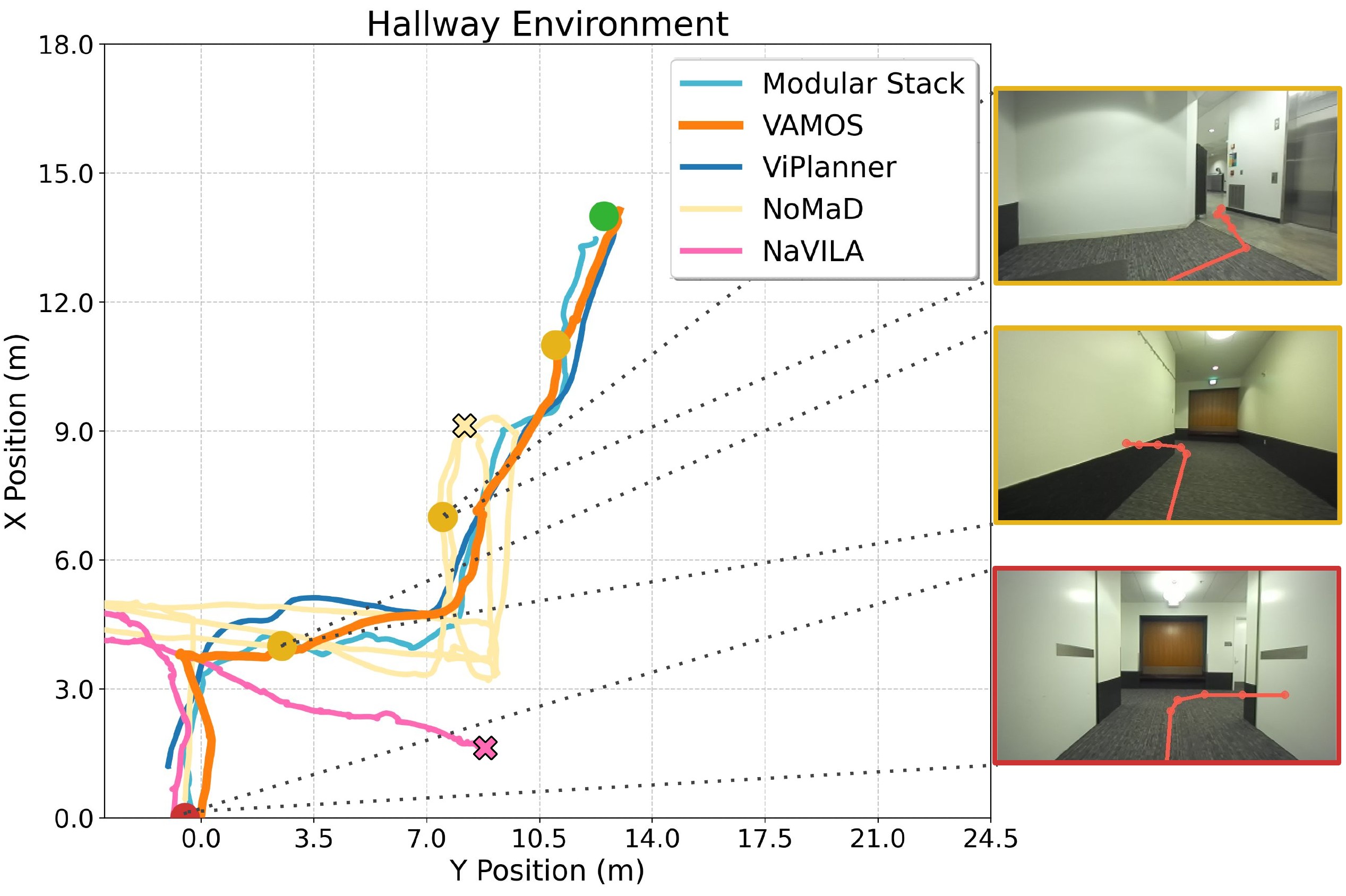}
        \caption{Hallway}
        \label{fig:hallway_topdown}
    \end{subfigure}
    \hfill
    \begin{subfigure}[b]{0.48\textwidth}
        \centering
        \includegraphics[height=5.625cm, keepaspectratio]{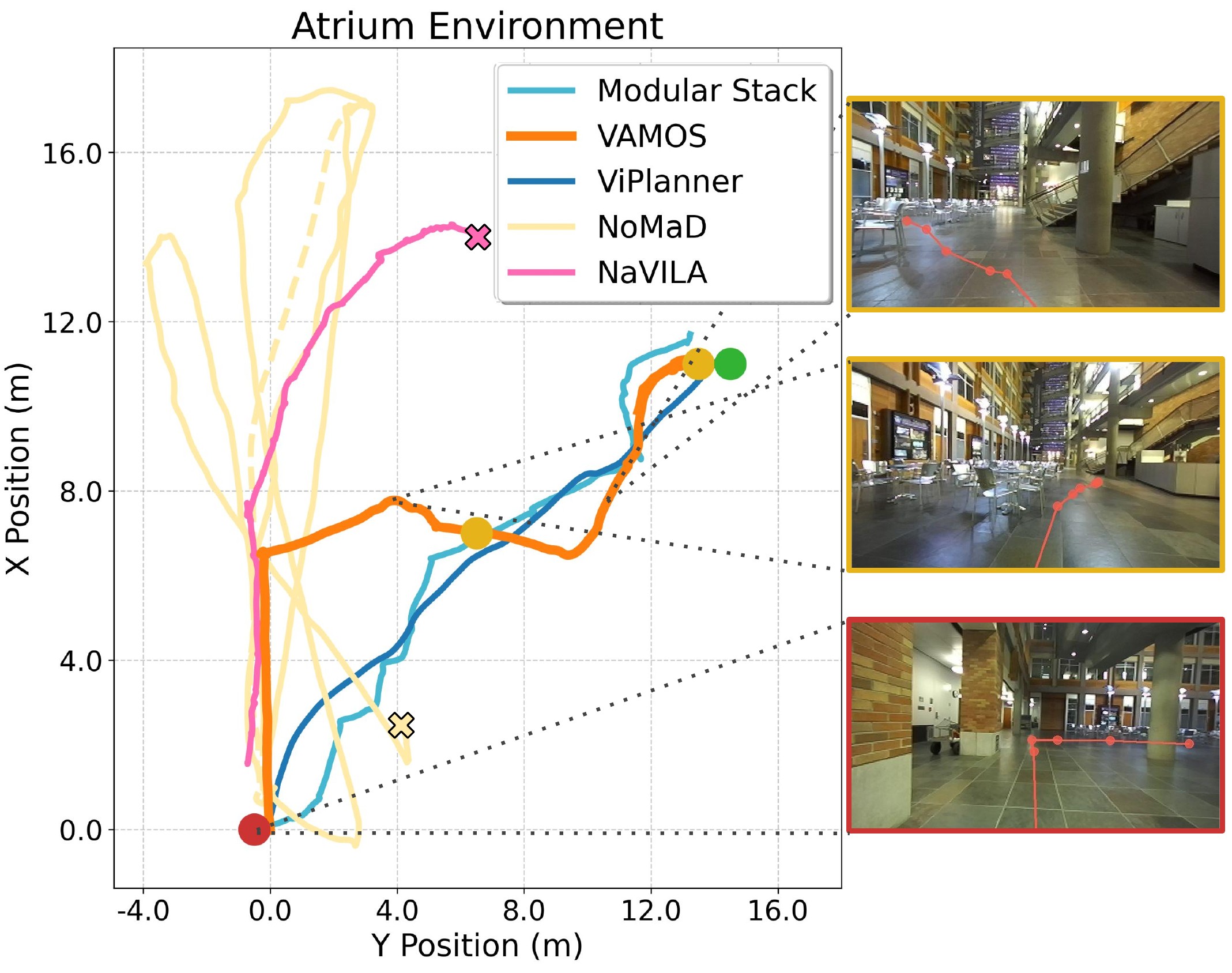}
        \caption{Atrium}
        \label{fig:atrium_topdown}
    \end{subfigure}

    \vspace{6pt}
    \begin{subfigure}[b]{0.48\textwidth}
        \centering
        \includegraphics[height=5.625cm, keepaspectratio]{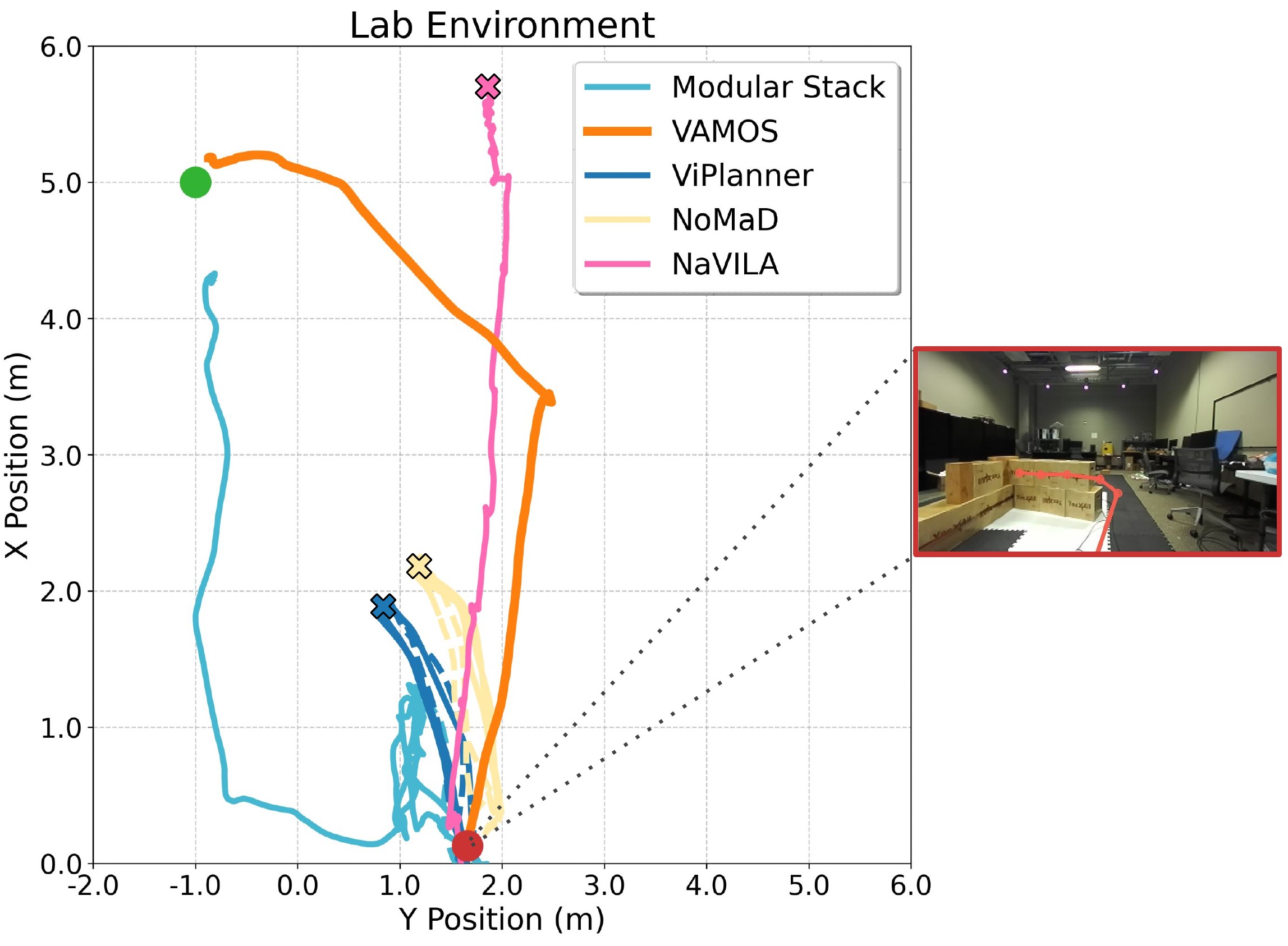}
        \caption{Lab}
        \label{fig:lab_topdown}
    \end{subfigure}
    \hfill
    \begin{subfigure}[b]{0.48\textwidth}
        \centering
        \includegraphics[height=5.625cm, keepaspectratio]{media/campus_top_down.jpg}
        \caption{Campus}
        \label{fig:campus_topdown}
    \end{subfigure}

    \vspace{6pt}
    \begin{subfigure}[b]{0.48\textwidth}
        \centering
        \includegraphics[height=5.625cm, keepaspectratio]{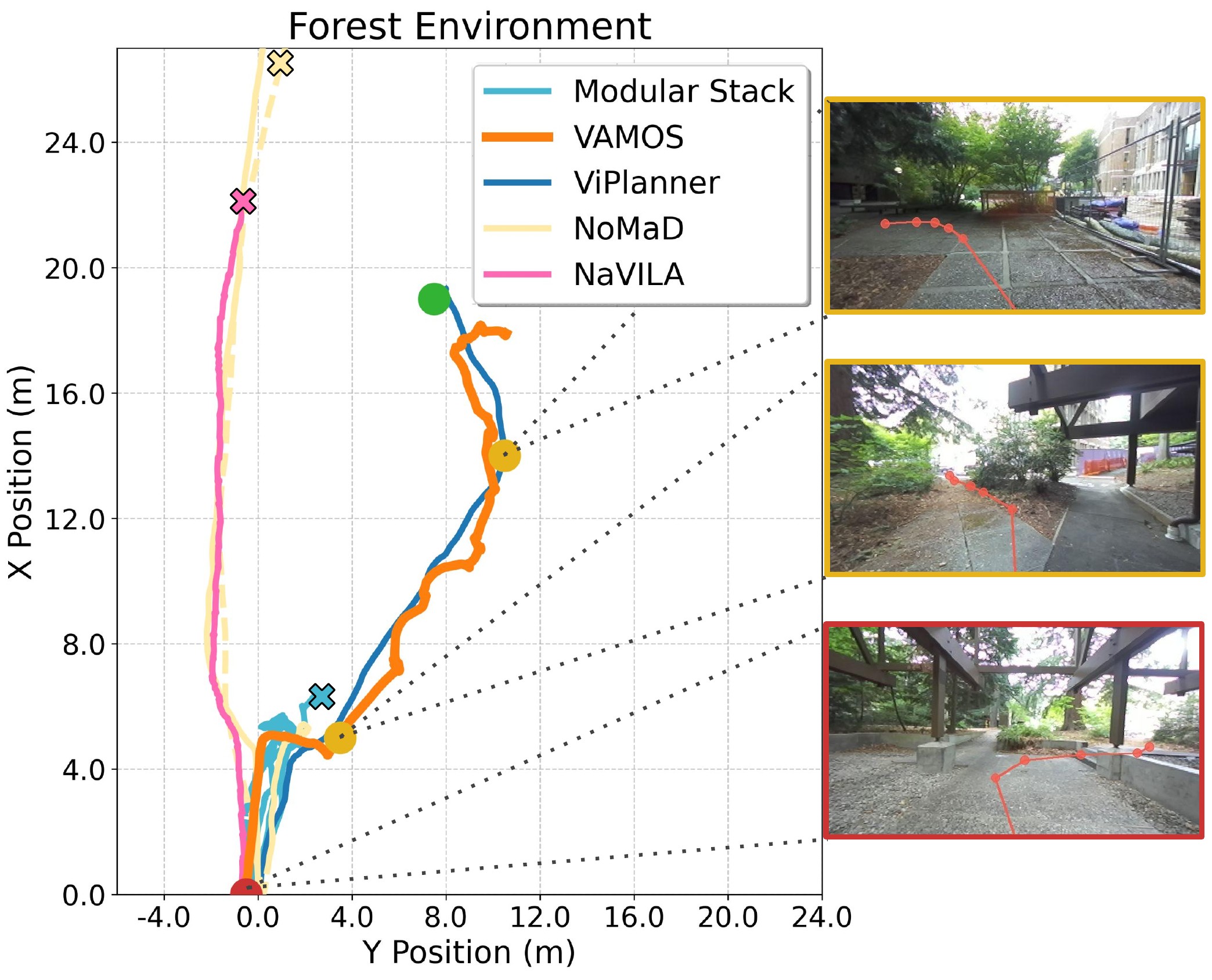}
        \caption{Forest}
        \label{fig:forest_topdown}
    \end{subfigure}
    \hfill
    \begin{subfigure}[b]{0.48\textwidth}
        \centering
        \includegraphics[height=5.625cm, keepaspectratio]{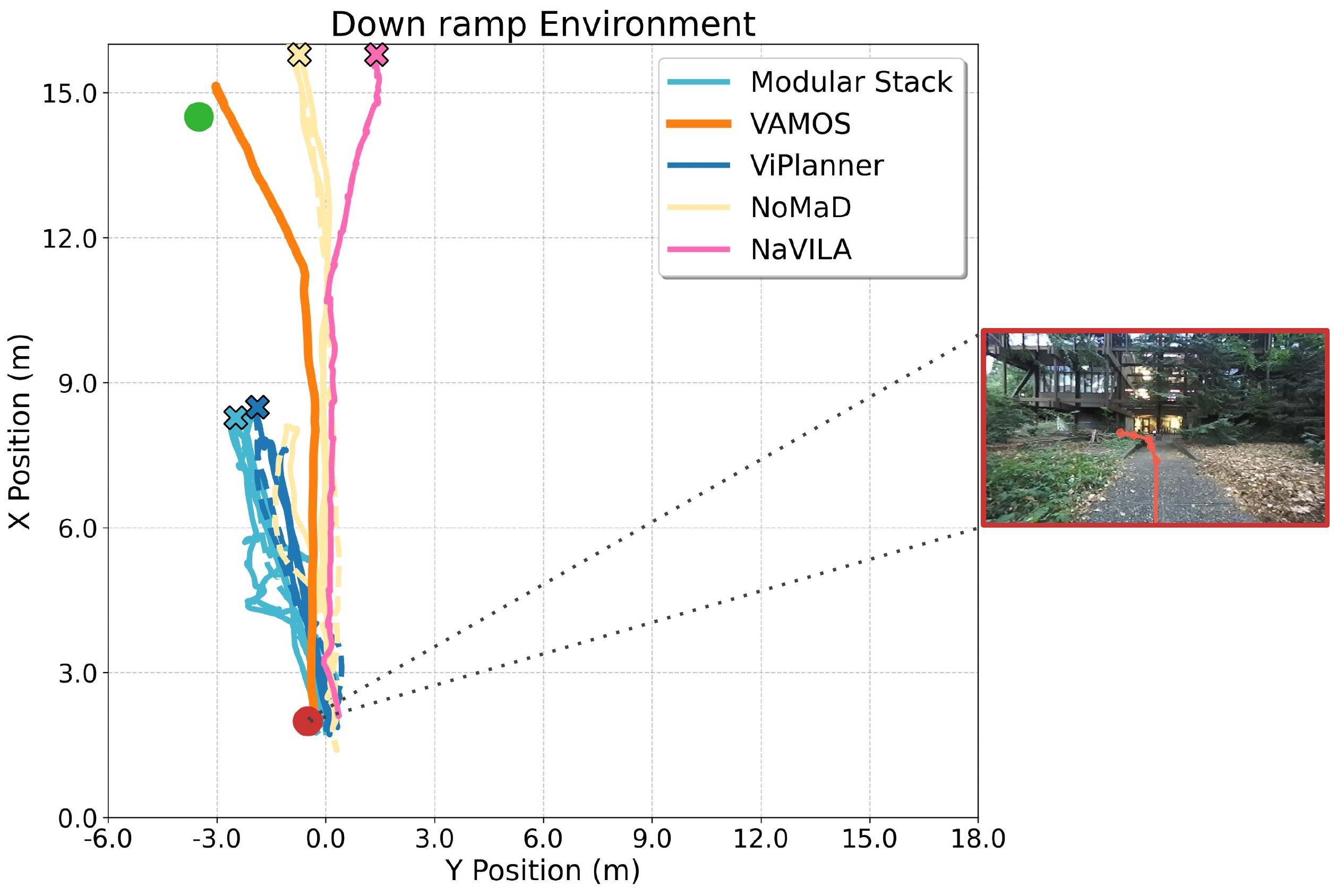}
        \caption{Down Ramp}
        \label{fig:downramp_topdown}
    \end{subfigure}

    \caption{We show a top-down map for all navigation courses of the paths taken by different methods to navigate from the start of the course (red circle) to the goal (green circle), through each waypoint (yellow circles). \methodname~is capable of long-horizon, precise navigation. To the right, we visualize the paths predicted and selected by \methodname~when replanning after reaching a waypoint. Dotted lines correspond to taking the robot back to the last previously-completed waypoint after interventions, and X's correspond to the positions where baselines failed or timed-out.}
    \label{fig:all_courses_topdown}
\end{figure*}

\begin{figure*}[htb]  %
  \centering
  \begin{subfigure}[t]{0.48\textwidth}
    \includegraphics[width=\linewidth,height=0.25\textheight,keepaspectratio]{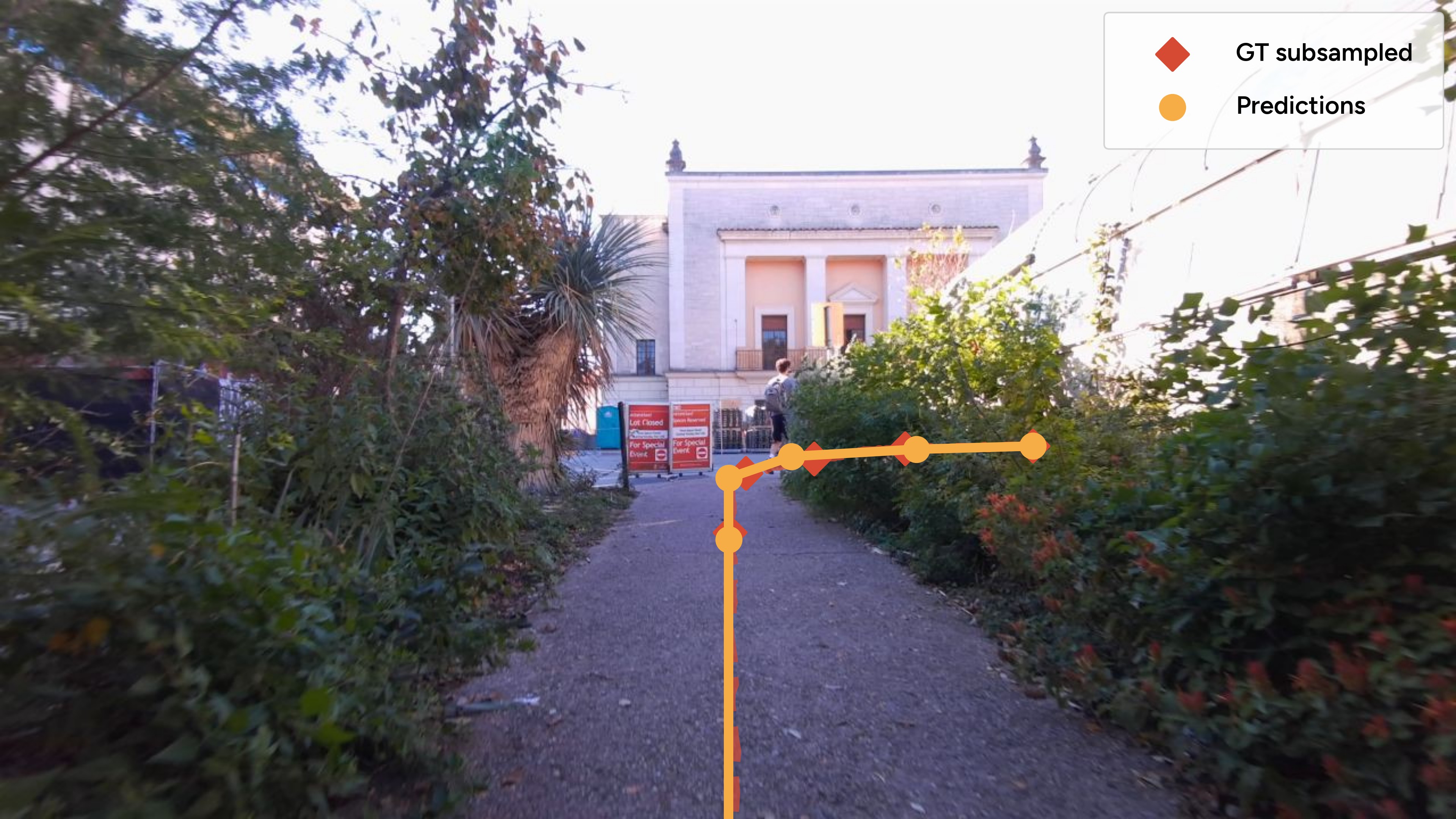}
    \caption{SCAND}
    \label{fig:scand_good_sample}
  \end{subfigure}
  \hfill
  \begin{subfigure}[t]{0.48\textwidth}
    \includegraphics[width=\linewidth,height=0.25\textheight,keepaspectratio]{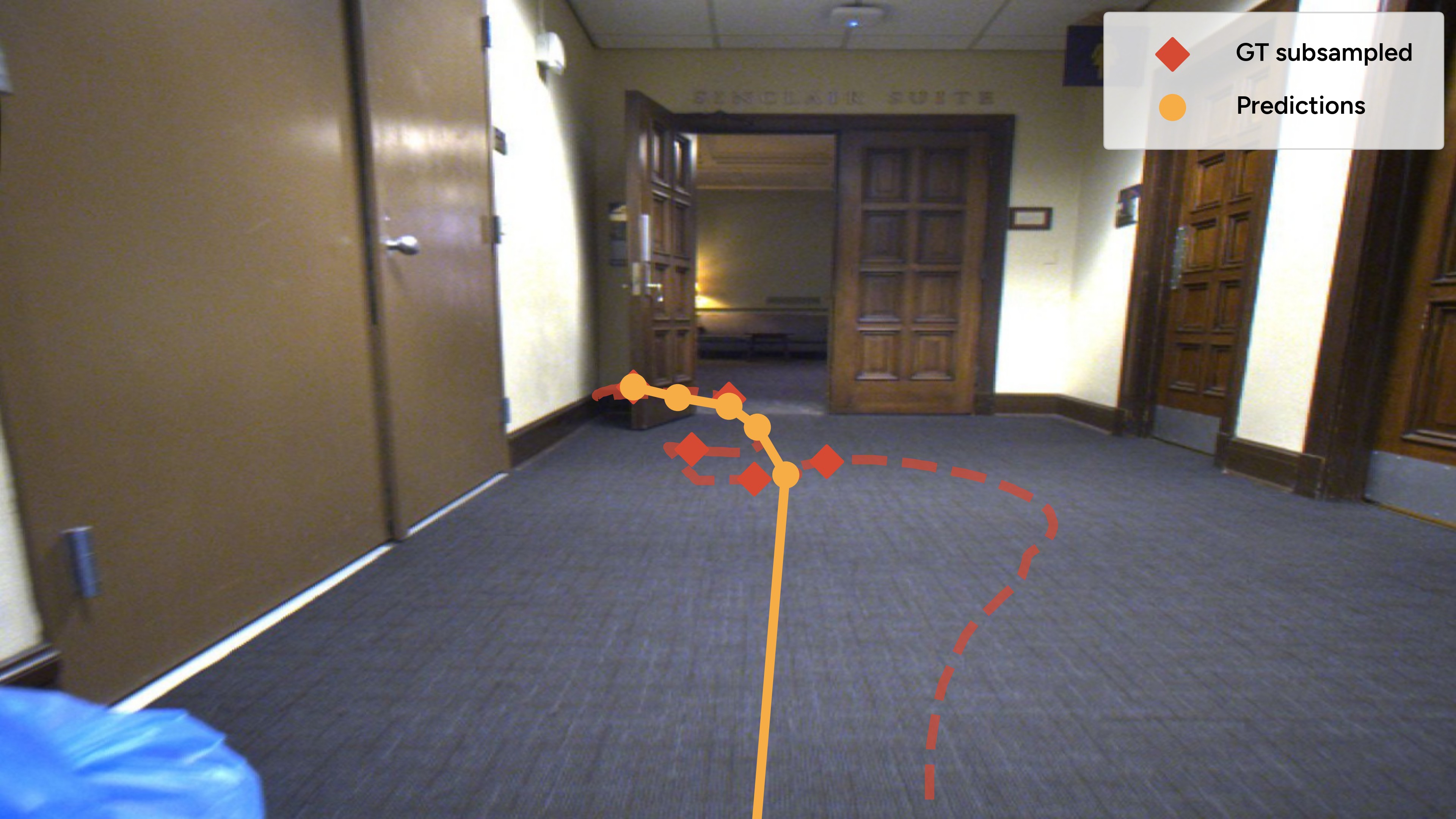}
    \caption{CODa}
    \label{fig:coda_good_sample}
  \end{subfigure}

  \par\bigskip

  \begin{subfigure}[t]{0.48\textwidth}
    \includegraphics[width=\linewidth,height=0.25\textheight,keepaspectratio]{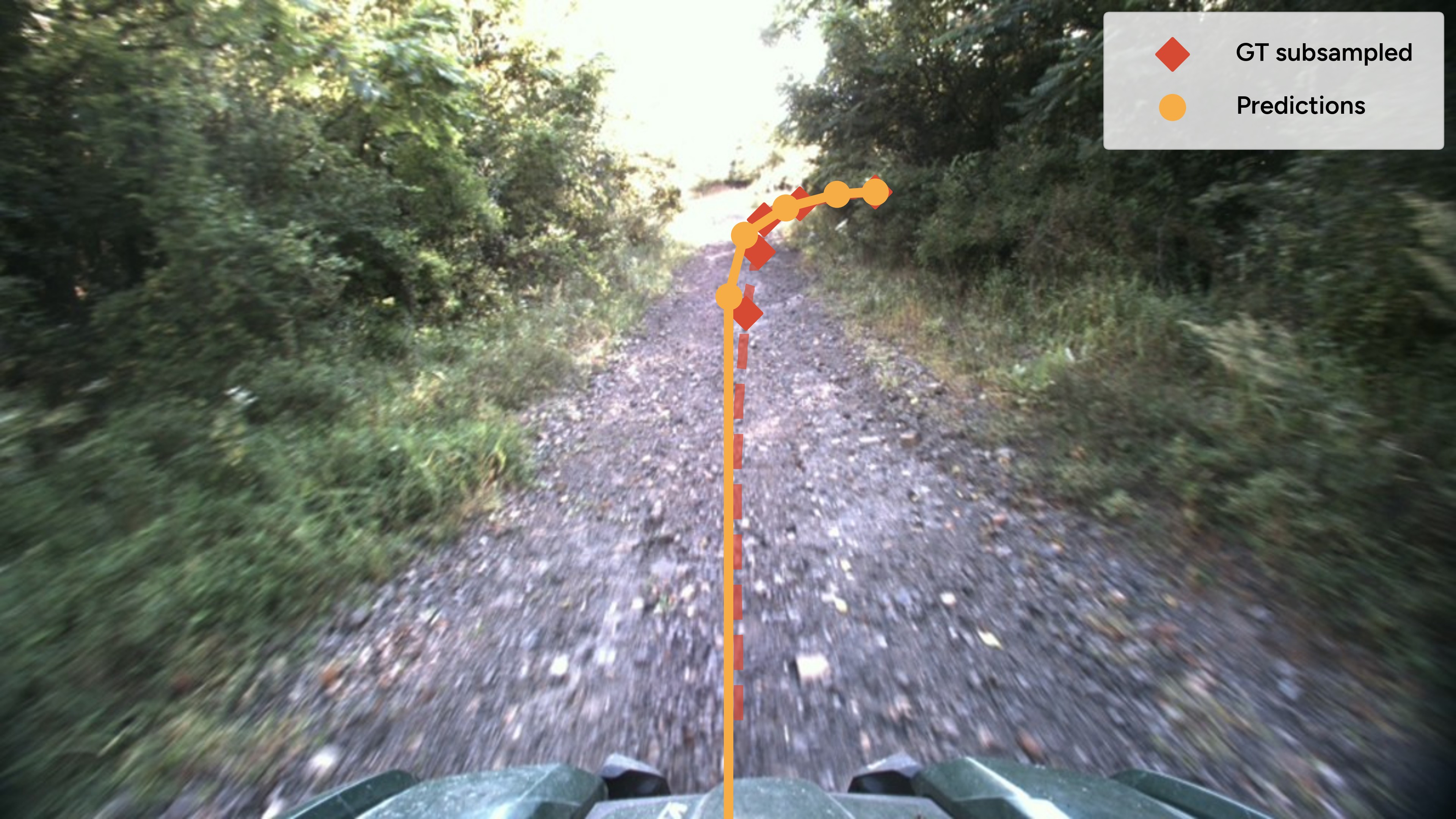}
    \caption{TartanDrive}
    \label{fig:tartandrive_good_sample}
  \end{subfigure}
  \hfill
  \begin{subfigure}[t]{0.48\textwidth}
    \includegraphics[width=\linewidth,height=0.25\textheight,keepaspectratio]{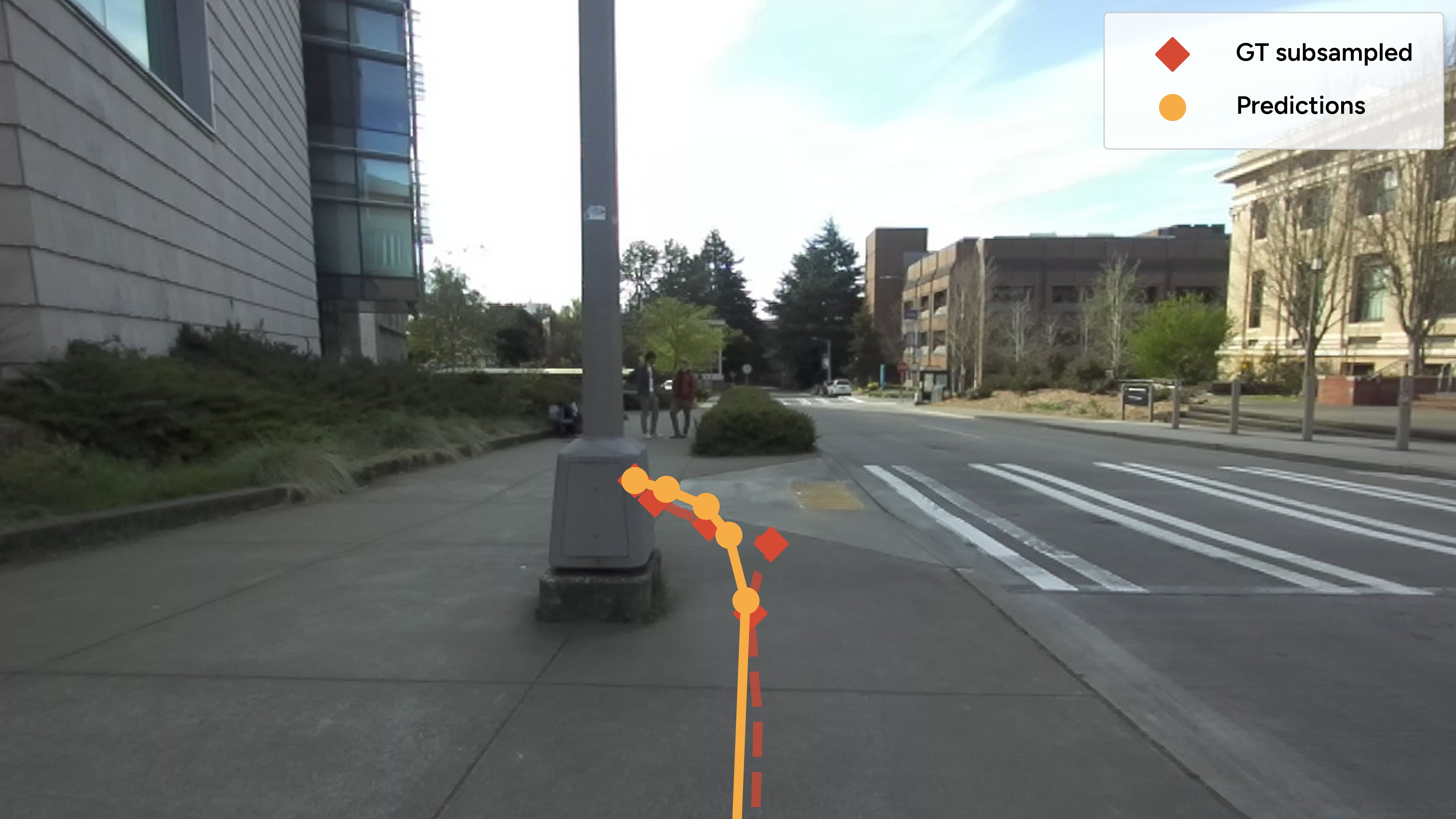}
    \caption{Spot}
    \label{fig:spot_good_sample}
  \end{subfigure}

  \caption{\textbf{Examples of high-level trajectory predictor.} The high-level navigator consistently gets to the goal, is good at following paths, going around obstacles, and taking turns behind occlusions such as walls, people, poles, etc.}
  \label{fig:good_samples}
\end{figure*}

\begin{figure*}[htb]  %
  \centering
  \begin{subfigure}[t]{0.48\textwidth}
    \includegraphics[width=\linewidth,height=0.25\textheight,keepaspectratio]{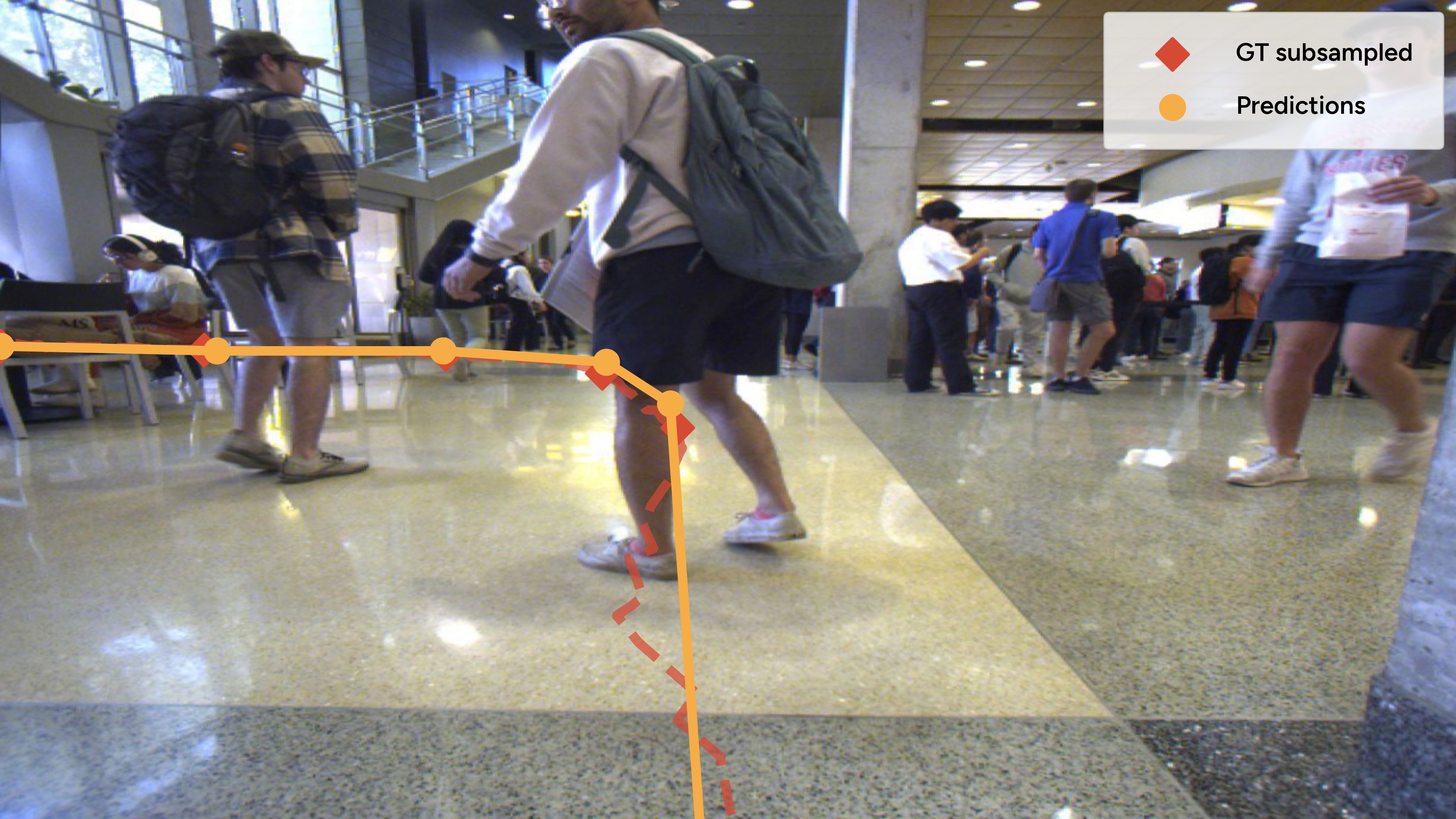}
    \caption{Dynamic Obstacles}
    \label{fig:dynamic_obstacles}
  \end{subfigure}
  \hfill
  \begin{subfigure}[t]{0.48\textwidth}
    \includegraphics[width=\linewidth,height=0.25\textheight,keepaspectratio]{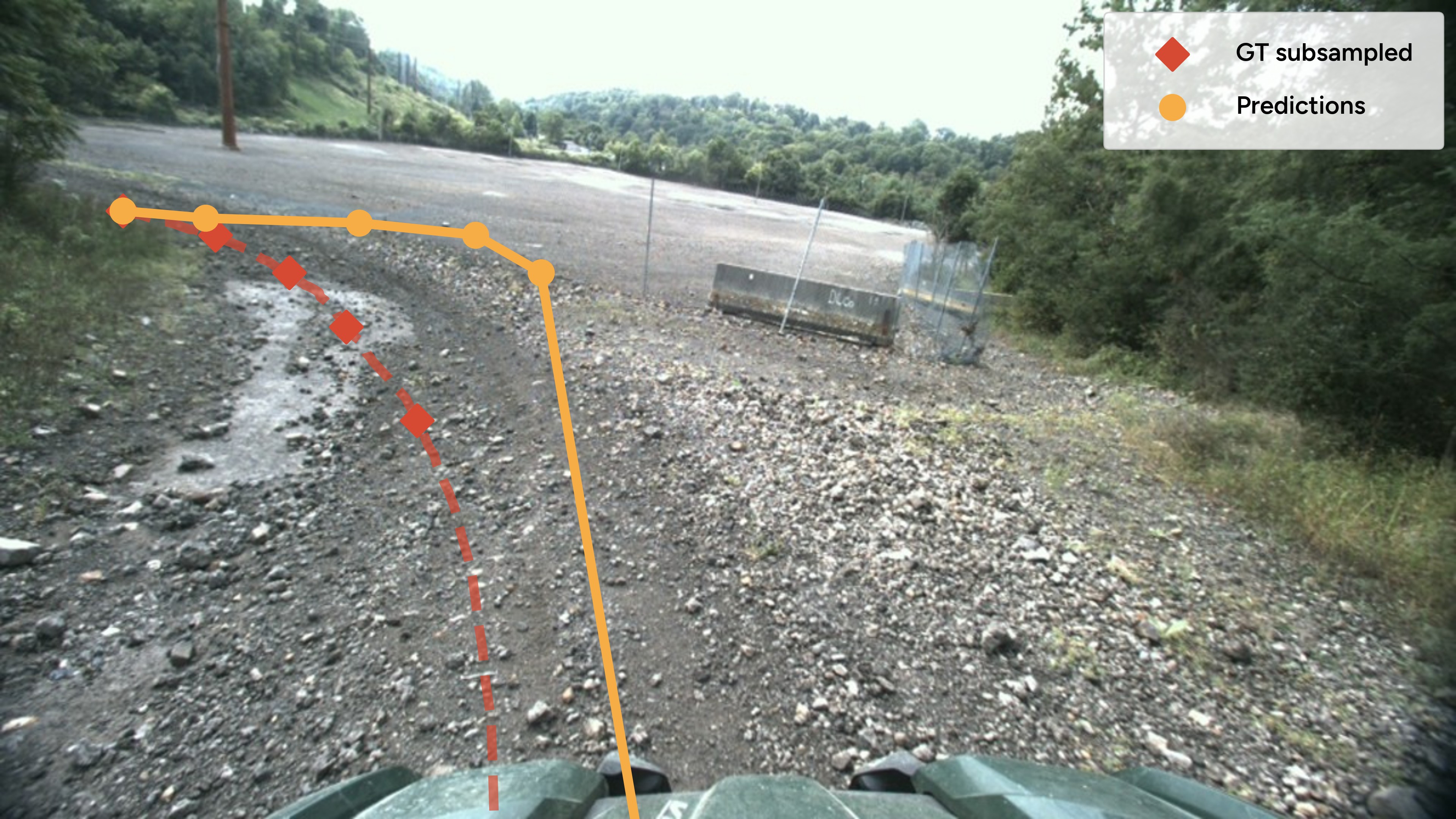}
    \caption{Overshooting}
    \label{fig:overshooting}
  \end{subfigure}

  \caption{\textbf{Examples of failure modes.} The high-level navigator sometimes struggles with dynamic obstacles, as in the training data dynamic obstacles usually move out of the way and their motion is not captured in the training data. It also sometimes overshoots or undershoots turns.}
  \label{fig:failure_modes}
\end{figure*}

We show a visualization and top-down map ofall real-world navigation courses in Figure \ref{fig:all_courses_topdown}. We also show some examples of the predicted and ground truth trajectories in each dataset in Figure \ref{fig:good_samples}. Our model is good at following paths and trails, going behind obstacles and occlusions, and reaching the goal. Given that the ground-truth trajectories are long-horizon, sometimes these paths take roundabout ways to reach the goal. Our high-level VLM often takes direct paths to the goal. We show some of the VLM's failure modes in Figure \ref{fig:failure_modes}. Two salient failure modes are dynamic obstacles and over/under-shooting turns. Given the staticness of the training data, the model does not capture close-by dynamic obstacles, such as walking people, very well. Additionally, the model sometimes overshoots or undershoots turns behind obstacles and occlusions. Sometimes this occurs due to the way in which we subsample trajectories uniformly, causing clipping at important points of the trajectories. However, our experiments with other subsampling methods that aim to capture salient points, such as the Ramer-Douglas-Peucker algorithm \cite{ramer1972iterative, douglas1973algorithms} (as is done in \cite{li2025hamster}) showed that this type of sampling hurt performance.

\section{Traversability Function Details}
\begin{figure}[!h]
    \centering
    \includegraphics[width=0.7\linewidth]{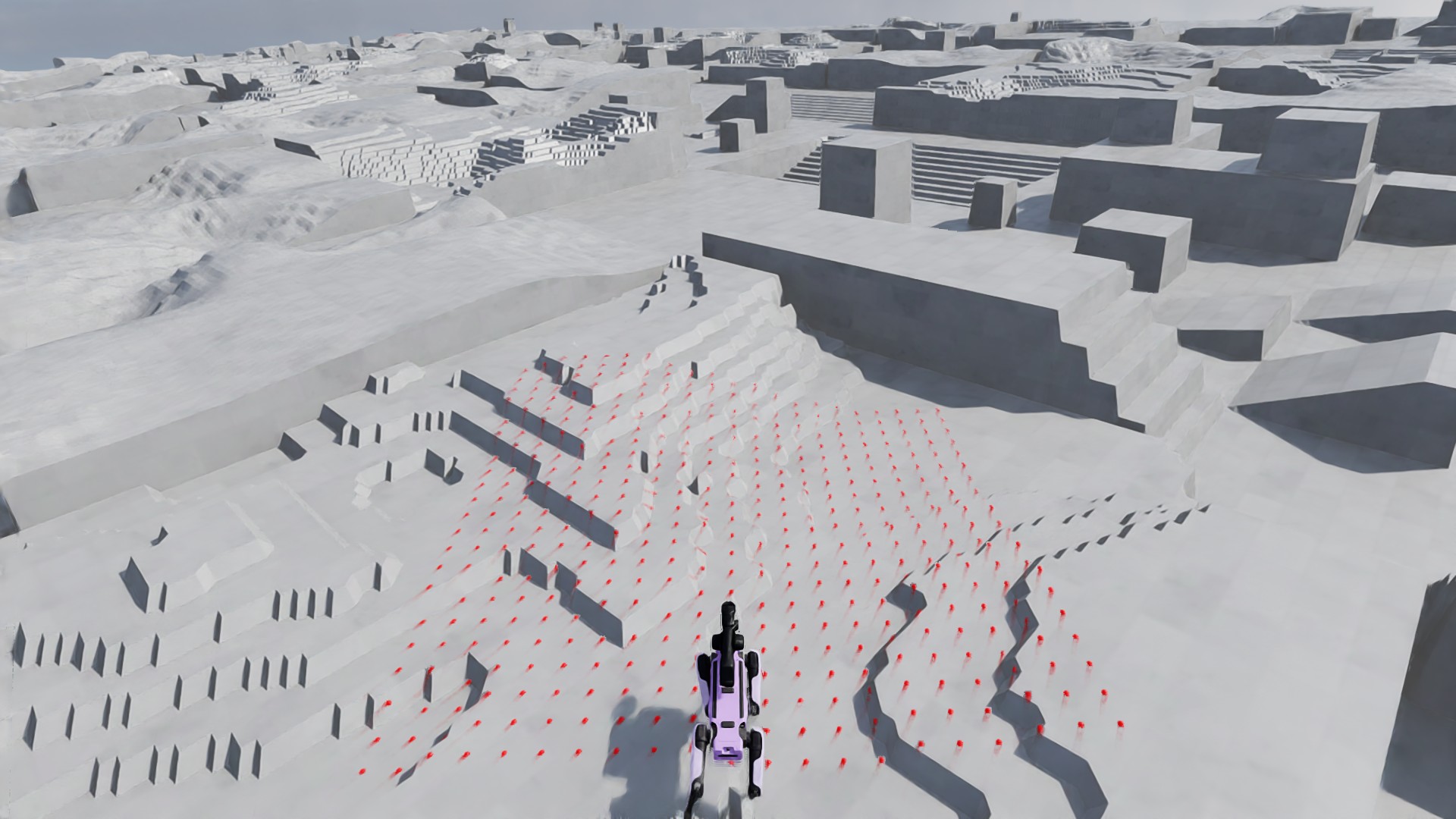}
    \caption{We generate terrains in simulation to train the affordance function using a combination of procedurally generated stairs- and ramp-like terrains with different parameterizations.}
    \label{fig:terrain-generation}
\end{figure}
\subsection{Terrain Generation}
For adequate sim to real transfer we found that it was important to generate a varied set of terrains(Figure \ref{fig:terrain-generation}) to simulate the diversity of the real world. We chose to use 5 different terrain types: irregular stairs, smooth mounds, procedurally generates stair and ramp environments, simple ramp, and simple stairs.\\
\textbf{Simple Stairs: } For the simple stairs terrain we have a 2m long by 10m wide flat area on both sides of the terrain, then a 6m long by 10m wide staircase connecting the two flat areas. The step width is set to 0.4m and the step height is drawn from a uniform distribution from 0.05m to 0.15m. \\
\textbf{Simple Ramp: } The simple ramp is similar to the simple stairs except the two flat regions are connected by a ramp. The slope of this ramp is drawn from a uniform distribution from 0.01m to 0.3m. \\
\textbf{Procedural Terrain: } The procedural terrain is composed of 25 two by two square tiles. Each tile in the terrain can either be a box, ramp, stairs, or flat. Then we use wave function collapse to populate all of the tiles and ensure they adhere to certain rules. We want stairs to either connect to other stairs or a flat area, and we want the area at the top of the stairs to be at the same height as the top of the stairs. Additionally, we randomize the heights of each stair or ramp and the sizes and heights of the boxes.\\
\textbf{Smooth Mounds}: To generate smooth mounds we use cellular automaton. We first choose $n$ random cells in our height map to serve as seeds. Each of these positions is set to some height drawn from a uniform distribution between 1m and 3m. Then we set the value of every cell in the heightmap to the value of the closest seed. Finally, to smooth everything out, for each cell in the height map if the difference between the minimum and maximum neighbor is greater than some threshold, then we set the height at the cell to the mean of those two neighbors. In practice, we have found that this algorithm is able to generate irregular terrains similar to uneven ground outdoors.\\
\textbf{Irregular Stairs:} To generate a irregular terraced pattern, we first use the cellular automaton to generate the smooth mounds. Then given some step height, we round the heights of each cell to the nearest whole number multiple of the step height. This terrain is meant to make our value function more robust to sharp local changes in elevation.
\subsection{Dataset Generation}
We first generate 1000 different 10m by 10m terrains using the methods outlined above, with all of the terrain types being equally represented. Then we depending on the robot type we have slightly different methods for data collection.

\textbf{Spot: } For spot we select a uniformly random unit vector as our velocity target. Then we roll out the policy until it terminates or times out. We terminate the roll-outs when the robot hits the wall which we compute by Equation~\ref{eq:wall} where $v_r$ is the robot velocity vector, $v_c$ is the command velocity vector, and $\tau$ is some threshold.
\begin{equation}
    \mathbf{1}\left\{ \frac{v_r \cdot v_c}{\|v_c\|} < \tau \right\}
    \label{eq:wall}
\end{equation}
In practice, we use $\tau=0.3$. Additionally, we only compute this termination after the first $0.5$ seconds to allow the robot to initially accelerate. The other termination that we use is a penalty for falling when either the velocity in the $z$ direction is less than $-1$ or the robot is tilted more than $45^\circ$ degrees on the roll or pitch axes. The rewards for each timestep correspond to the terminations. We have a reward of $-1$ when we terminate.

The policy we roll out in simulation is trained with PPO \cite{schulman2017proximal} in rough terrains using Isaac Lab \cite{mittal2023orbit} with proprioceptive and perceptive observations consisting of geometric height samples, following a terrain curriculum, similar to \cite{rudin2022learning}. Even though this is a different policy than the built-in Spot locomotion policy we use during deployment, it acts as a good surrogate for learning the capabilities of a performant all-terrain navigation policy.

\textbf{Hound: } For Hound we collect all trajectories by driving in a straight line because the affordances of the car over a small distance tend to be the same while driving strait and turning. We use the same terminations and rewards as Spot.

Instead of using a learned or default low-level controller, we use a pure-pursuit based controller with a kinematic bicycle model to reach various waypoints.

\subsection{Qualitative Comparison to Regression-based Value Function}

\begin{figure}[t]
    \centering
    \begin{subfigure}[b]{0.48\columnwidth}
        \centering
        \includegraphics[width=\linewidth]{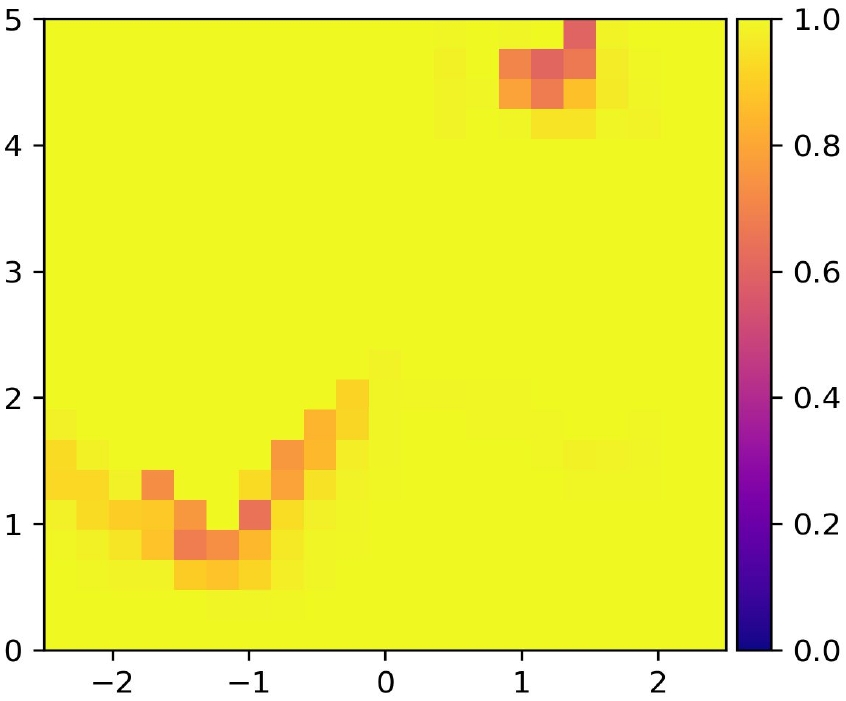}\\
        \includegraphics[width=\linewidth]{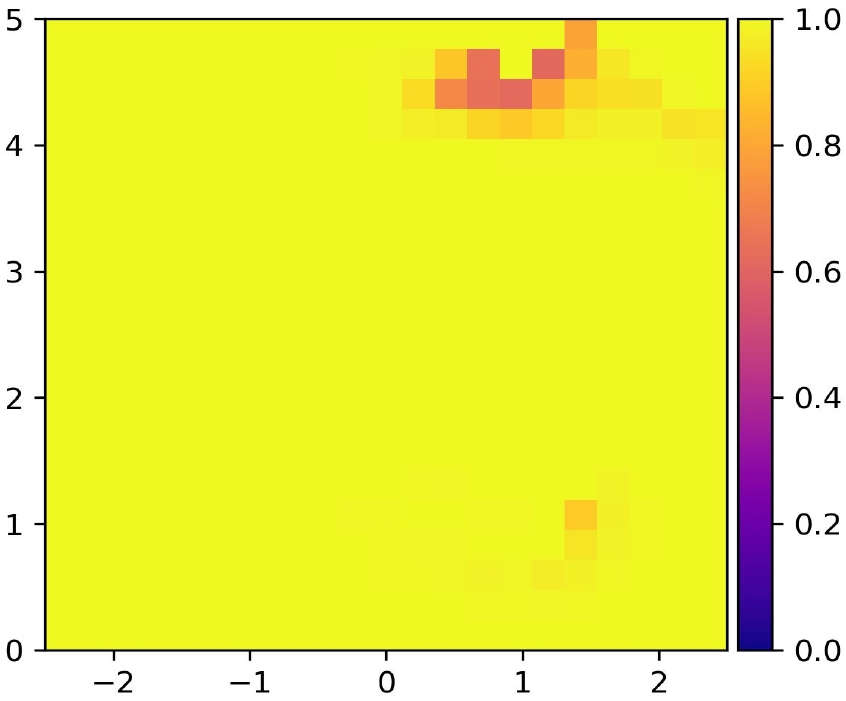}
        \caption{Regression on Returns}
        \label{fig:return_regression}
    \end{subfigure}
    \hfill
    \begin{subfigure}[b]{0.48\columnwidth}
        \centering
        \includegraphics[width=\linewidth]{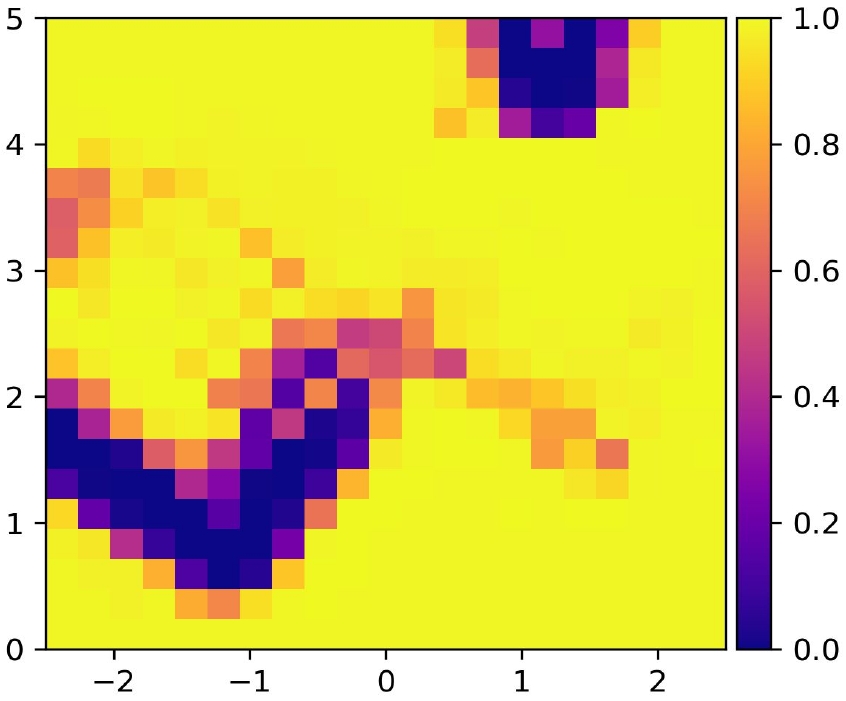}\\
        \includegraphics[width=\linewidth]{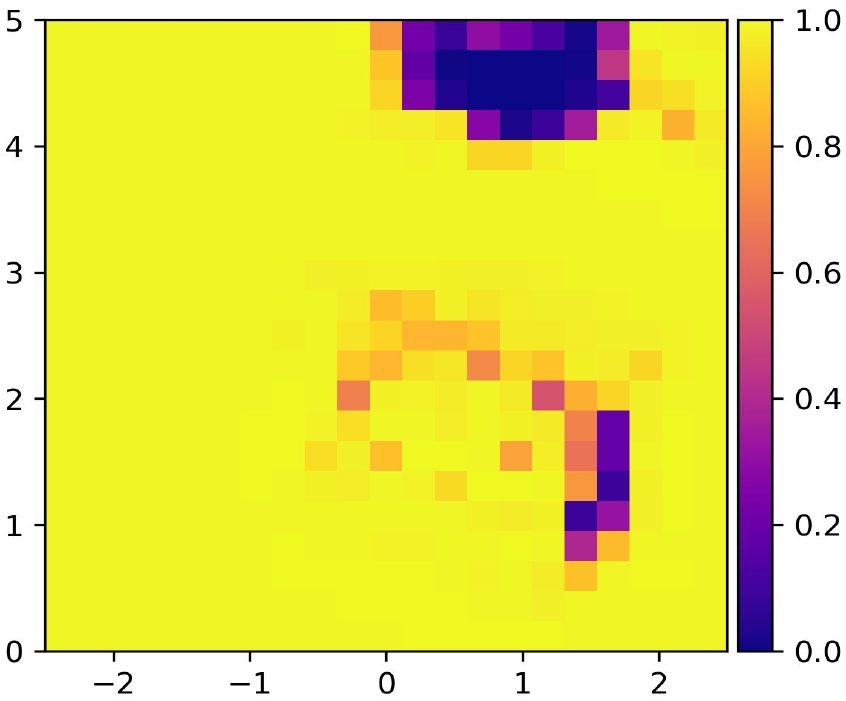}
        \caption{Classification on Labels}
        \label{fig:label_classification}
    \end{subfigure}

    \caption{Computing the affordance function as a classification task rather than a regression on returns, as is common in reinforcement learning, yields more discriminative affordance scores. Here, we show two examples from two different scenes, where each row represents a scene.}
    \label{fig:regression_vf}
\end{figure}

Rather than training the model to classify local elevation maps as failures or successes, we can compute the discounted sum of rewards for each rollout and train the model to regress this value given the local elevation map (see Figure \ref{fig:regression_vf}). However, in practice we observe that classifying faliure or success of a trajectory works better than regressing reward to go. We believe that the reason for this is that we can easily balance the classification dataset to include an equal proportion of successes and failures. Additionally, we believe the classification problem better represents the task because avoids coupling the labels of unrelated observations.
\section{Deployment}
\subsection{Compute and Sensors}

We deploy our fine-tuned PaliGemma 2 high-level navigation generalist on an external laptop with an Nvidia RTX 3080 Laptop GPU, while the low-level low-level traversability function runs onboard a Jetson Orin AGX. High-level inference runs at 1 Hz when the laptop is connected to external power, and around 0.5 Hz otherwise. For Spot experiments, the laptop is connected to the robot's network through ethernet for increased reliabilty. For HOUND experiments, the laptop is connected to the robot through a 5 GHz Wi-Fi hotspot running on the laptop. We found 5GHz Wi-Fi to provide much better capacity and latency over 2.4 GHz Wi-Fi, albeit less reliable outdoors.

As sensor readings for the traversability function, we use an Ouster OS-1 LiDAR and Spot's built-in depth cameras on all sides of the robot to construct a square 16x16 meter elevation map using \cite{miki2022elevation}, from which we crop smaller local grids for the traversability function observations. We use a Zed2i camera for the Spot robot and a Realsense D455 camera for the HOUND robot. 

\subsection{VLM Sampling}

For the navigation experiments outlined in Section \ref{sec:real_world_nav}, we sample the VLM with $\texttt{temperature}=0.1$, $\texttt{num\_beams}=1$, and no $\texttt{top\_k}$ nor $\texttt{top\_p}$ sampling. For the traversability function experiments outlined in Section \ref{sec:modulation}, the only difference is that we sample the VLM with $\texttt{temperature}=1.0$ for the obstacle avoidance experiments and $\texttt{temperature}=0.3$ for the embodiment experiments.

\subsection{State Machine}

We deploy \methodname~ in the real world within a simple state machine. First, the robot either plans with the VLM and then tracks the first $m$ out of 5 predicted waypoints, or it rotates in place until the goal is within the image frame and then plans with the high-level VLM. Then, after either reaching the first $m$ waypoints, or after a timeout set to 20 seconds, whichever happens first, we repeat the loop and re-plan or rotate to put the goal within the image frame. We find that $m=3$ works well for shorter courses, and $m=4$ works well for longer courses. 

\section{Modular Stack Baseline Details}
\label{sec:modular_baseline}

For completeness, we provide a detailed description of our ``Modular stack'' baseline, which is highly performant and serves as a comparison point in our experiments. This baseline includes robust state estimation, global and local path planning, terrain analysis, and a strong low-level control module:

\begin{itemize}
    \item State Estimation: We use Spot's built-in visual odometry, a production-level odometry system deployed across all Spot robots.
    
    \item Traversability Analysis: The geometric costmap from Multi-Modal Elevation Mapping (MEM, \cite{erni2023mem}) is employed for terrain assessment. This is the same costmap utilized in prior works such as \cite{mattamala2024wild}.
    
    \item Global Planning: Using the MEM costmap, we employ ARA* \cite{likhachev2003ara}, an incremental, anytime variant of A*, for efficient global path planning.
    
    \item Local Planning: A pure-pursuit controller is used locally. This approach achieves performance comparable to MPPI with a kinematic bicycle model while being simpler and computationally cheaper.
    
    \item Low-Level Control: Spot's built-in RL-MPC locomotion controller handles low-level control, providing a robust, production-ready policy for navigating challenging terrains.
\end{itemize}

\section{Additional Quantitative Results}

\begin{figure*}[htb]  %
  \centering
  \begin{subfigure}[t]{0.48\textwidth}
    \includegraphics[width=\linewidth,height=0.25\textheight,keepaspectratio]{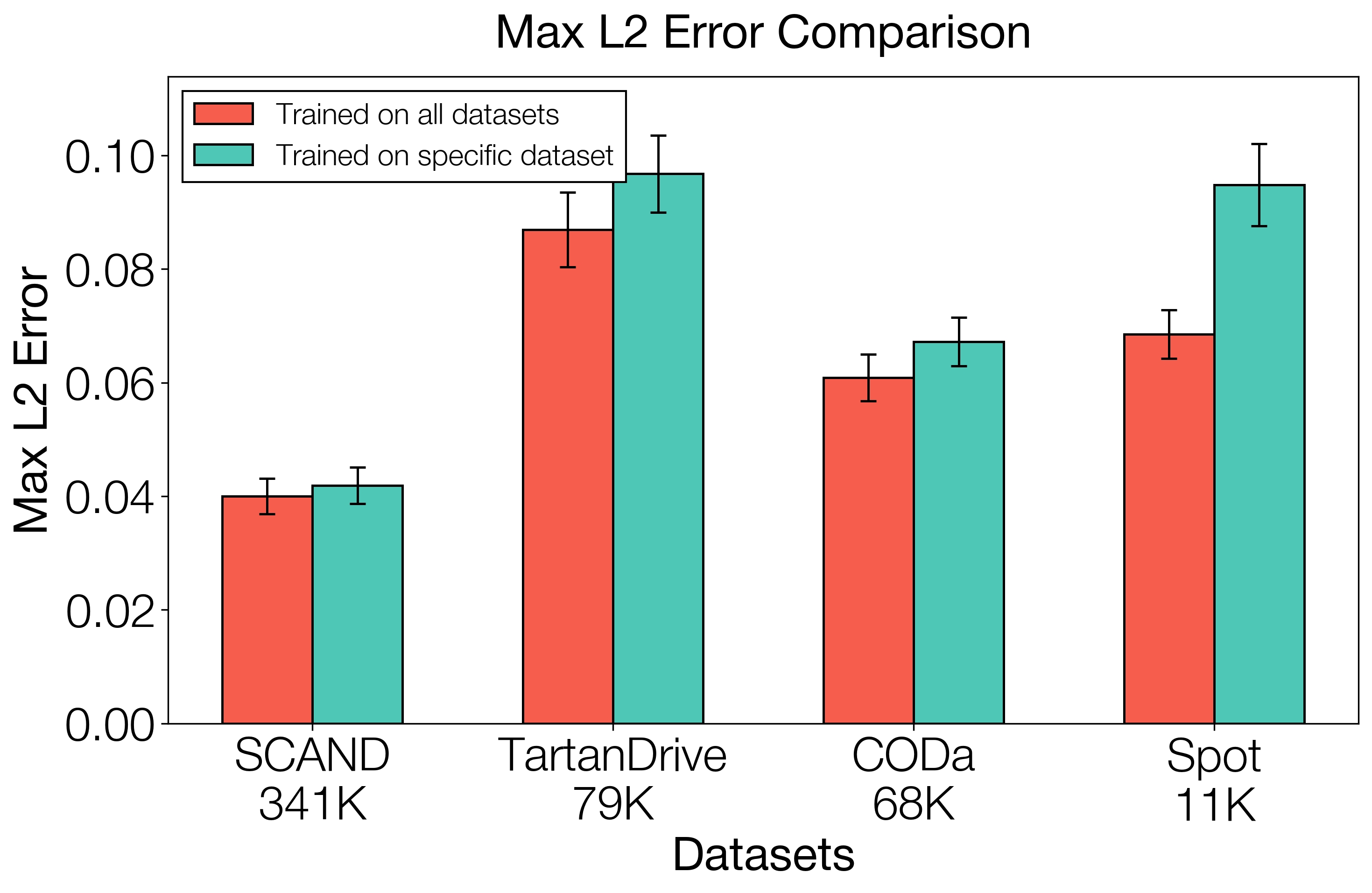}
    \caption{Max L2 Error}
    \label{fig:max_l2_error}
  \end{subfigure}
  \hfill
  \begin{subfigure}[t]{0.48\textwidth}
    \includegraphics[width=\linewidth,height=0.25\textheight,keepaspectratio]{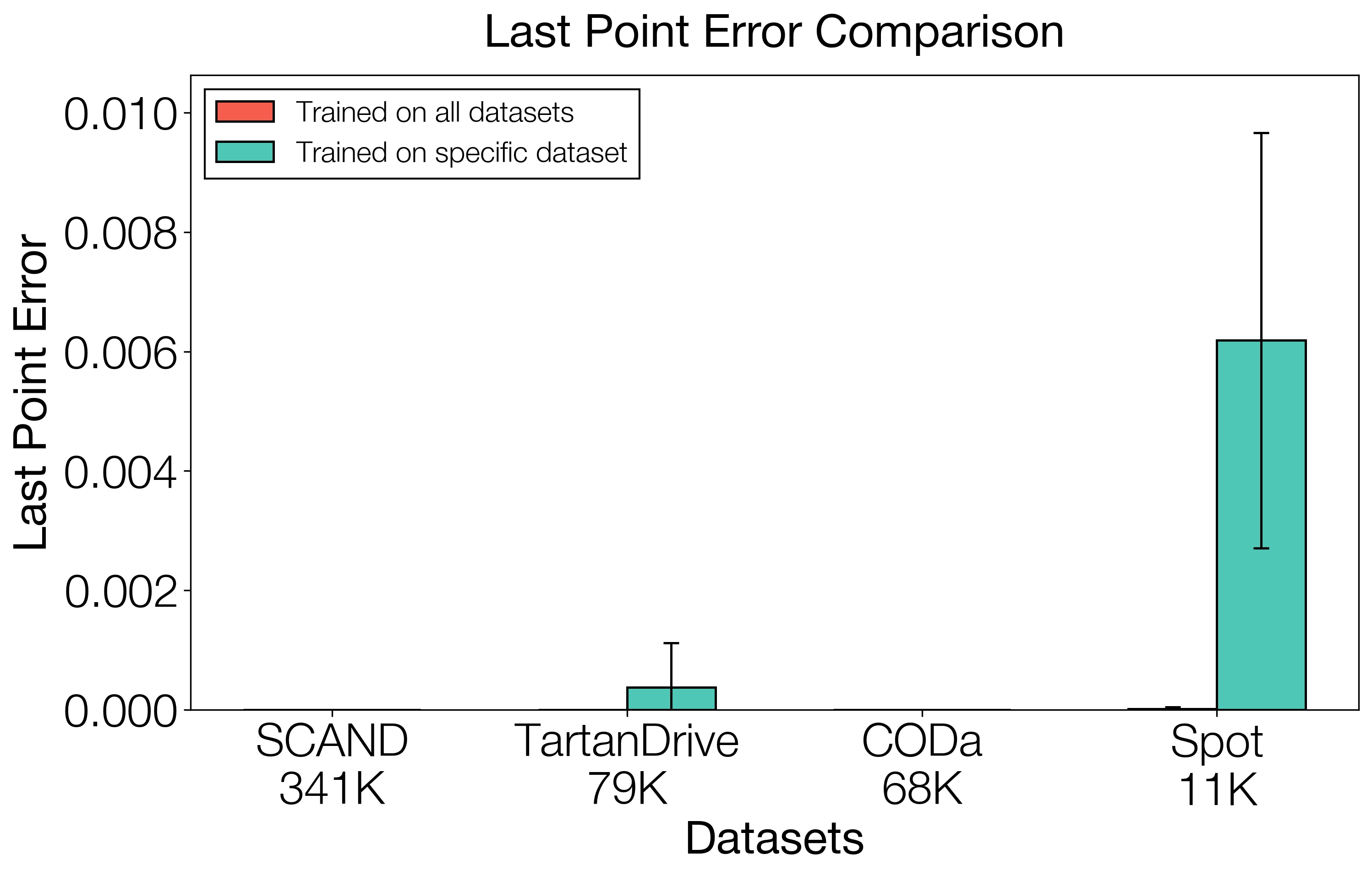}
    \caption{Last Point Error}
    \label{fig:last_point_error}
  \end{subfigure}

  \par\bigskip

  \begin{subfigure}[t]{0.48\textwidth}
    \includegraphics[width=\linewidth,height=0.25\textheight,keepaspectratio]{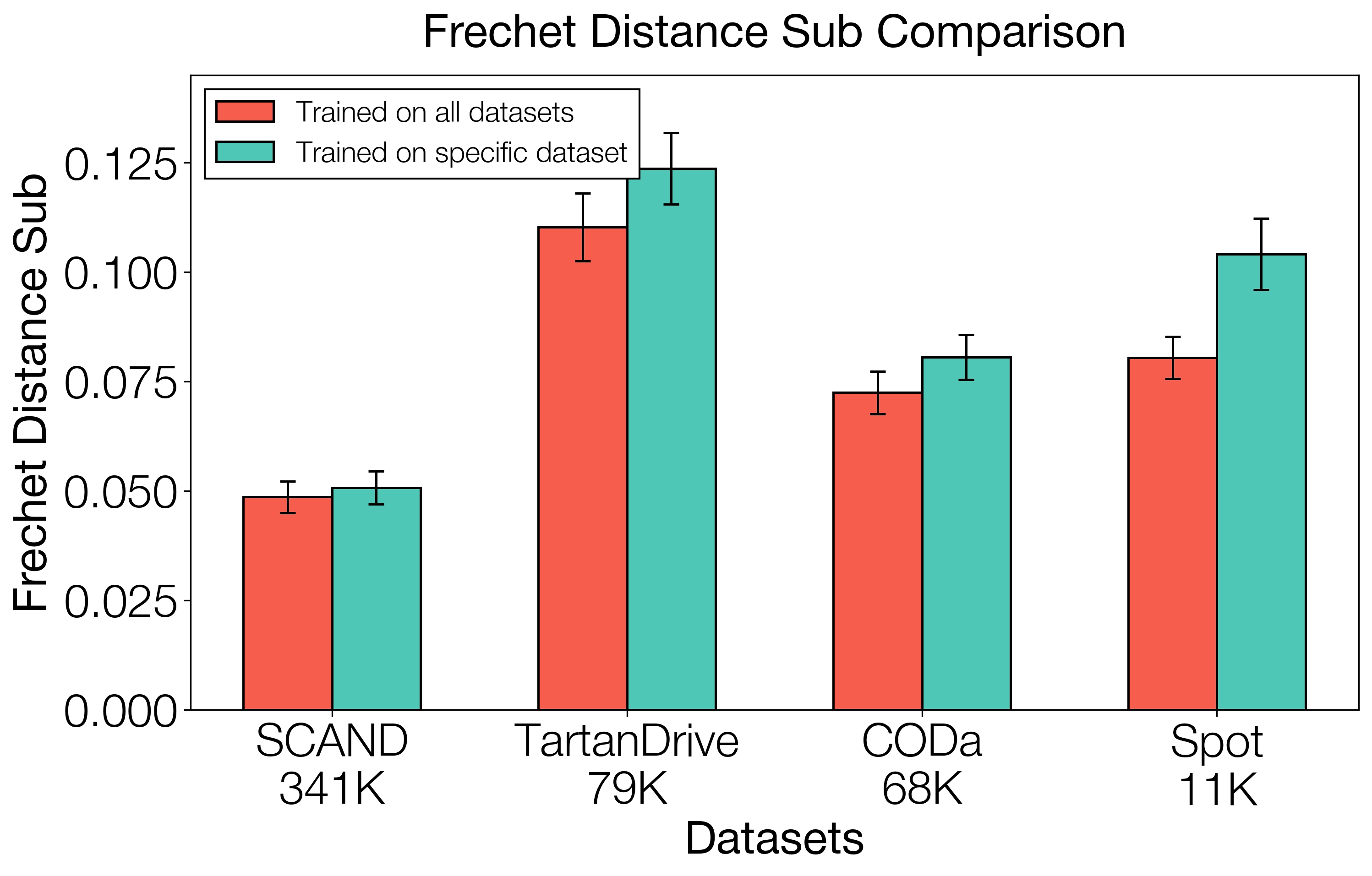}
    \caption{Fréchet Distance on Subsampled Trajectories}
    \label{fig:frechet_distance_sub}
  \end{subfigure}
  \hfill
  \begin{subfigure}[t]{0.48\textwidth}
    \includegraphics[width=\linewidth,height=0.25\textheight,keepaspectratio]{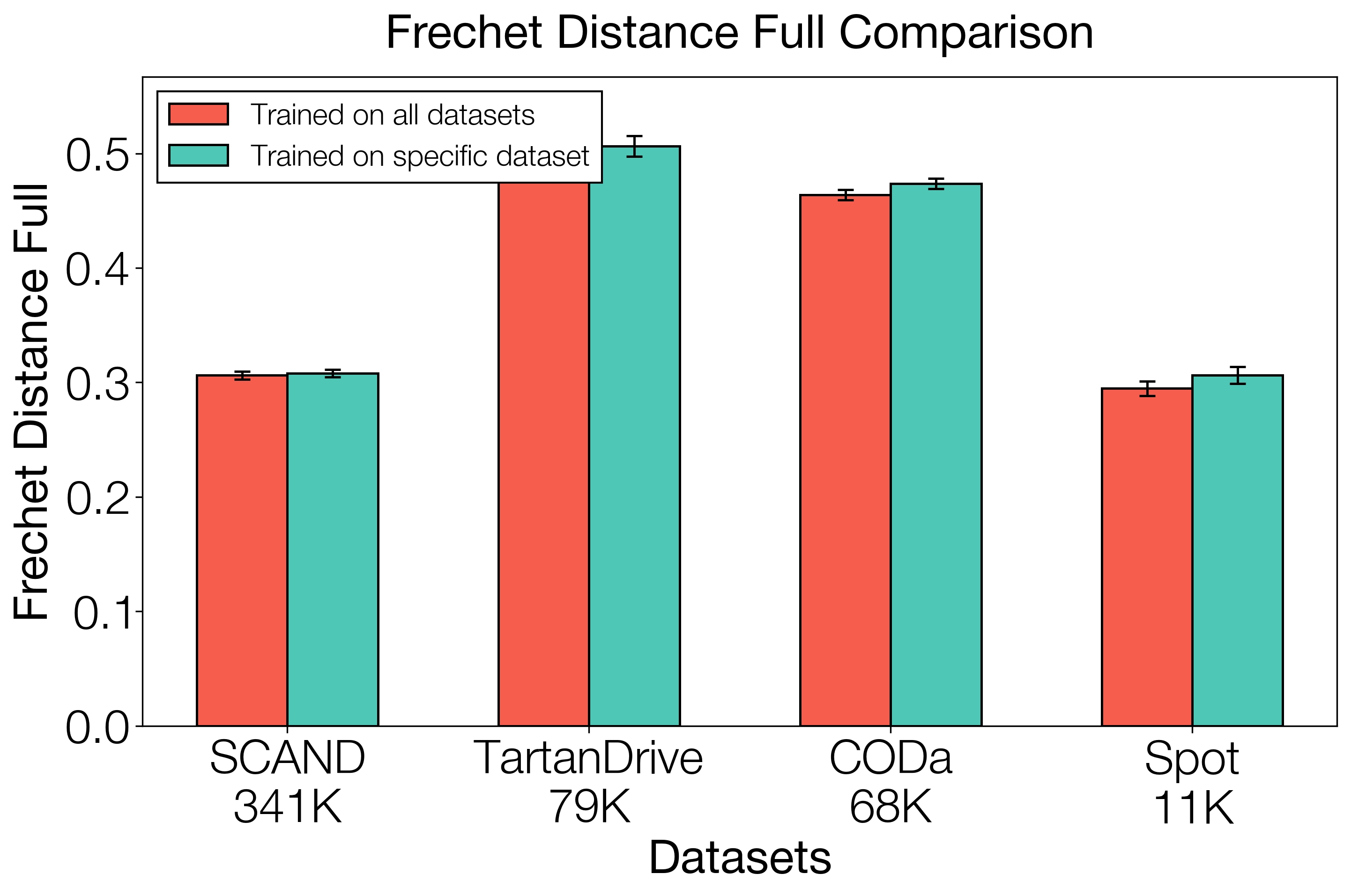}
    \caption{Fréchet Distance on Full Trajectories}
    \label{fig:frechet_distance_full}
  \end{subfigure}

  \par\bigskip

  \begin{subfigure}[t]{0.48\textwidth}
    \includegraphics[width=\linewidth,height=0.25\textheight,keepaspectratio]{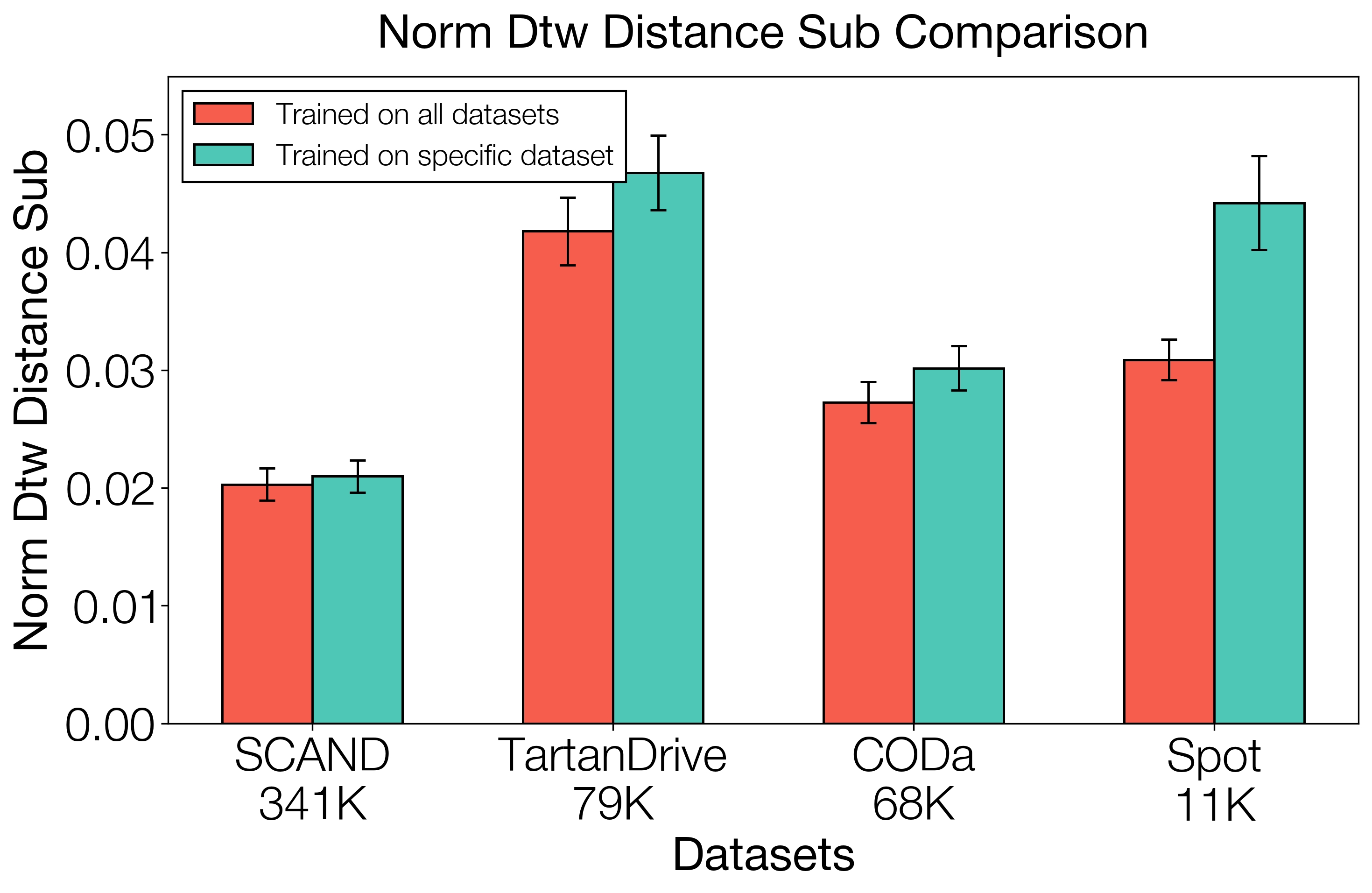}
    \caption{DTW Distance on Subsampled Trajectories}
    \label{fig:dtw_sub}
  \end{subfigure}
  \hfill
  \begin{subfigure}[t]{0.48\textwidth}
    \includegraphics[width=\linewidth,height=0.25\textheight,keepaspectratio]{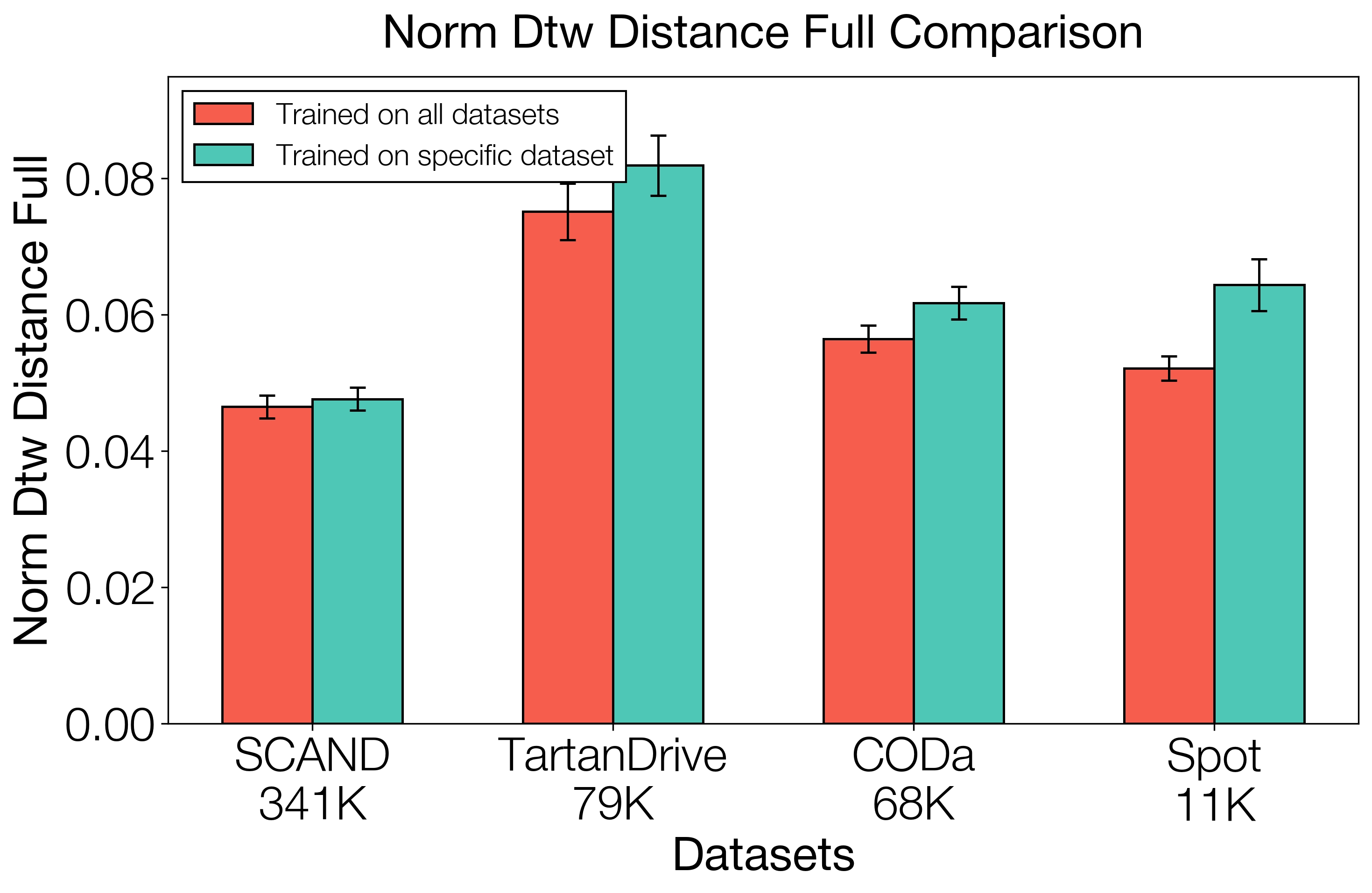}
    \caption{DTW Distance on Full Trajectories}
    \label{fig:dtw_full}
  \end{subfigure}

  \caption{Offline metrics comparing high-level generalist vs robot-specific navigators. In all metrics, we benefit from training on the pooled data compared to robot-specific datasets.}
  \label{fig:other_offline_metrics}
\end{figure*}

We compare our high-level generalist with robot-specific models using additional metrics to measure offline performance as  mentioned in Section \ref{sec:pooling}. We consider the following metrics, as shown in Figure \ref{fig:other_offline_metrics}:
\begin{itemize}
    \item \textbf{Mean L2 Error:} (Fig. \ref{fig:poolingviz}) Measures the error for all 5 points in a predicted trajectory, averaged across each trajectory, across all trajectories in the validation dataset.
    \item \textbf{Max L2 Error: } (Fig. \ref{fig:max_l2_error}) Measures the maximum error between all 5 points in a predicted trajectory, averaged across all trajectories in the validation dataset.
    \item \textbf{Fréchet Distance on Subsampled Trajectories: } (Fig. \ref{fig:frechet_distance_sub}) Measures the Fréchet distance between the subsampled ground-truth trajectory (i.e. the 5 points used as labels during training) and the predictions. Similar to a \texttt{max} function.
    \item \textbf{Fréchet Distance on Full Trajectories: } (Fig. \ref{fig:frechet_distance_full}) Measures the Fréchet distance between the full, dense, ground-truth trajectory (i.e. the original trajectories of hundreds of datapoints, depending on the horizon, subsampled at 10 Hz) and the 5-point predictions. Similar to a \texttt{max} function.
    \item \textbf{Dynamic Time Warping Distance on Subsampled Trajectories: } (Fig. \ref{fig:dtw_sub}) Measures the normalized Dynamic Time Warping distance between the subsampled ground-truth trajectory (i.e. the 5 points used as labels during training) and the predictions. Similar to a \texttt{mean} function.
    \item \textbf{Dynamic Time Warping Distance on Full Trajectories: } (Fig. \ref{fig:dtw_full}) Measures the normalized Dynamic Time Warping distance between the full, dense, ground-truth trajectory (i.e. the original trajectories of hundreds of datapoints, depending on the horizon, subsampled at 10 Hz) and the 5-point predictions. Similar to a \texttt{mean} function.
\end{itemize}

Additionally, we run statistical significance tests for all metrics on a per-dataset basis. Across these four datasets, the generalist model consistently yields small statistically significant improvements in trajectory accuracy. In TartanDrive, mean L2 and sub-sampled Fréchet distances improve at $p < 0.05$, while full-trajectory Fréchet and normalized DTW show highly significant gains (***). CODA exhibits highly significant reductions ($p < 10^{-5}$) in mean and max L2, sub-sampled Fréchet, and sub-sampled DTW, with endpoint and full-trajectory Fréchet errors remaining comparable. In SCAND, the general model outperforms on mean L2 (*), max L2 (**), sub-sampled Fréchet (**), and full-trajectory DTW (*), with other metrics not significant. On Spot, nearly all metrics except endpoint error and full-Fréchet errors achieve *** significance. Overall, these results suggest the general model produces smoother, more accurate paths across varied environments, even when terminal or peak deviations remain similar.

\begin{figure*}[htb]  %
  \centering
  \begin{subfigure}[t]{0.48\textwidth}
    \includegraphics[width=\linewidth,height=0.25\textheight,keepaspectratio]{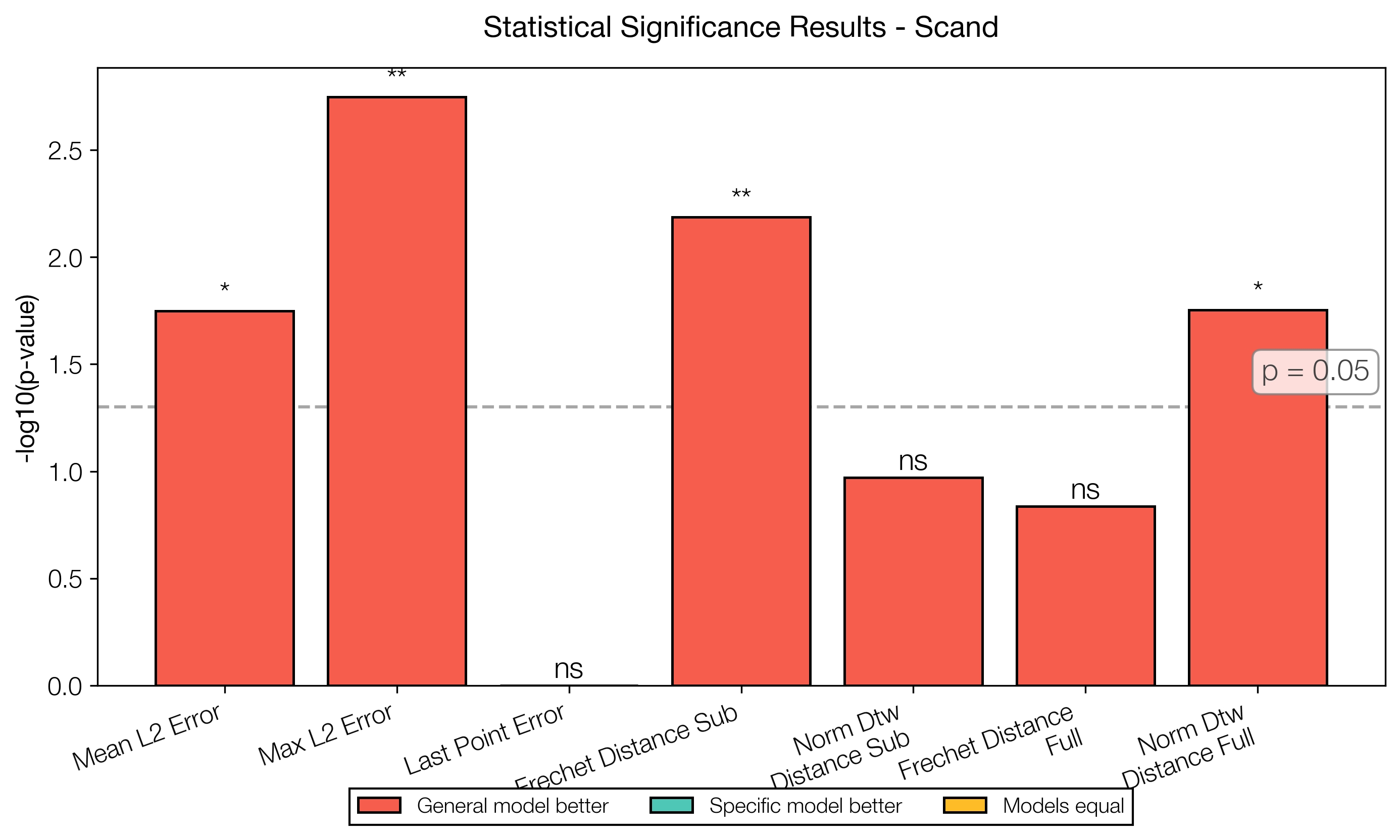}
    \caption{SCAND}
    \label{fig:scand_significance}
  \end{subfigure}
  \hfill
  \begin{subfigure}[t]{0.48\textwidth}
    \includegraphics[width=\linewidth,height=0.25\textheight,keepaspectratio]{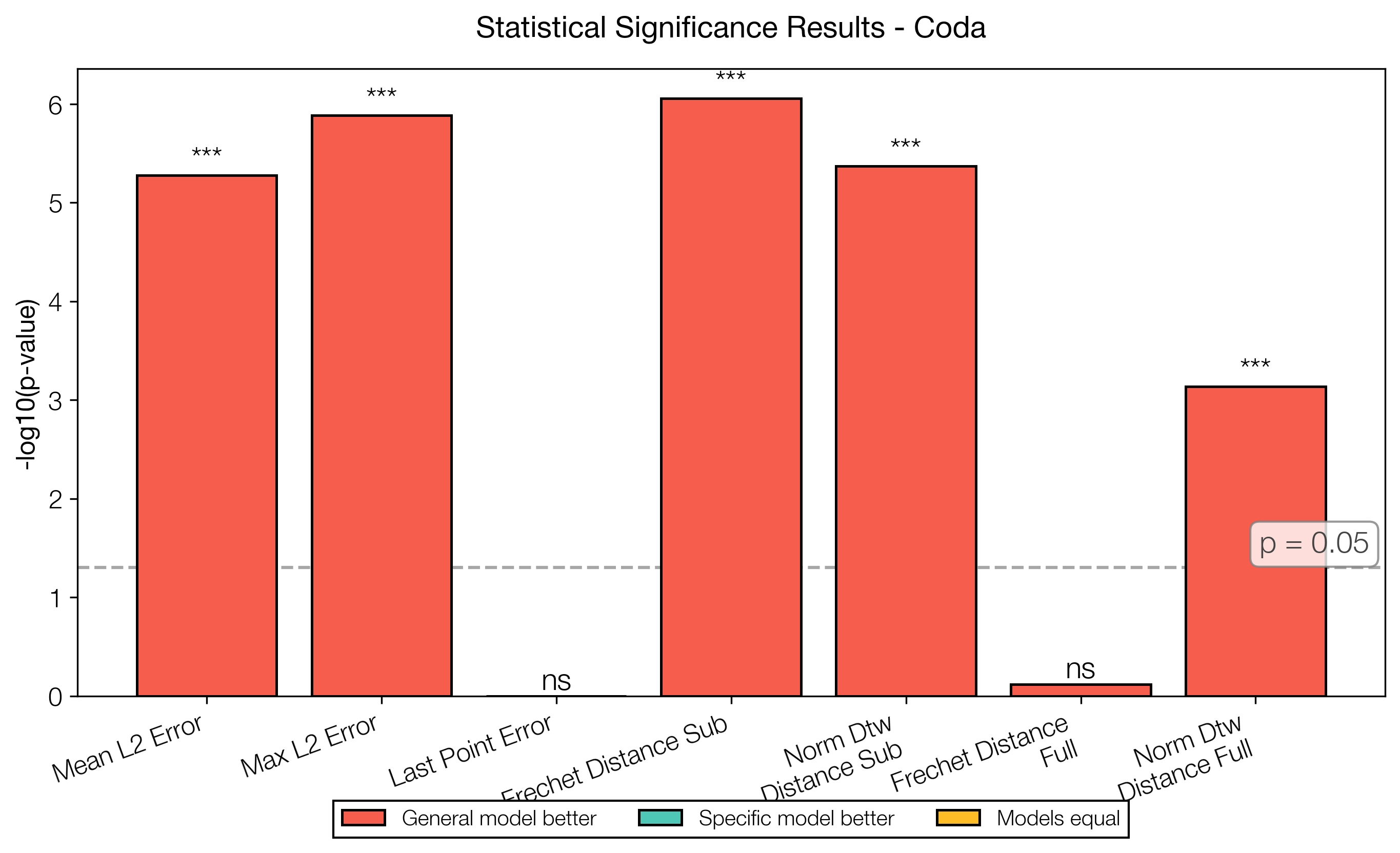}
    \caption{CODa}
    \label{fig:coda_significance}
  \end{subfigure}

  \par\bigskip

  \begin{subfigure}[t]{0.48\textwidth}
    \includegraphics[width=\linewidth,height=0.25\textheight,keepaspectratio]{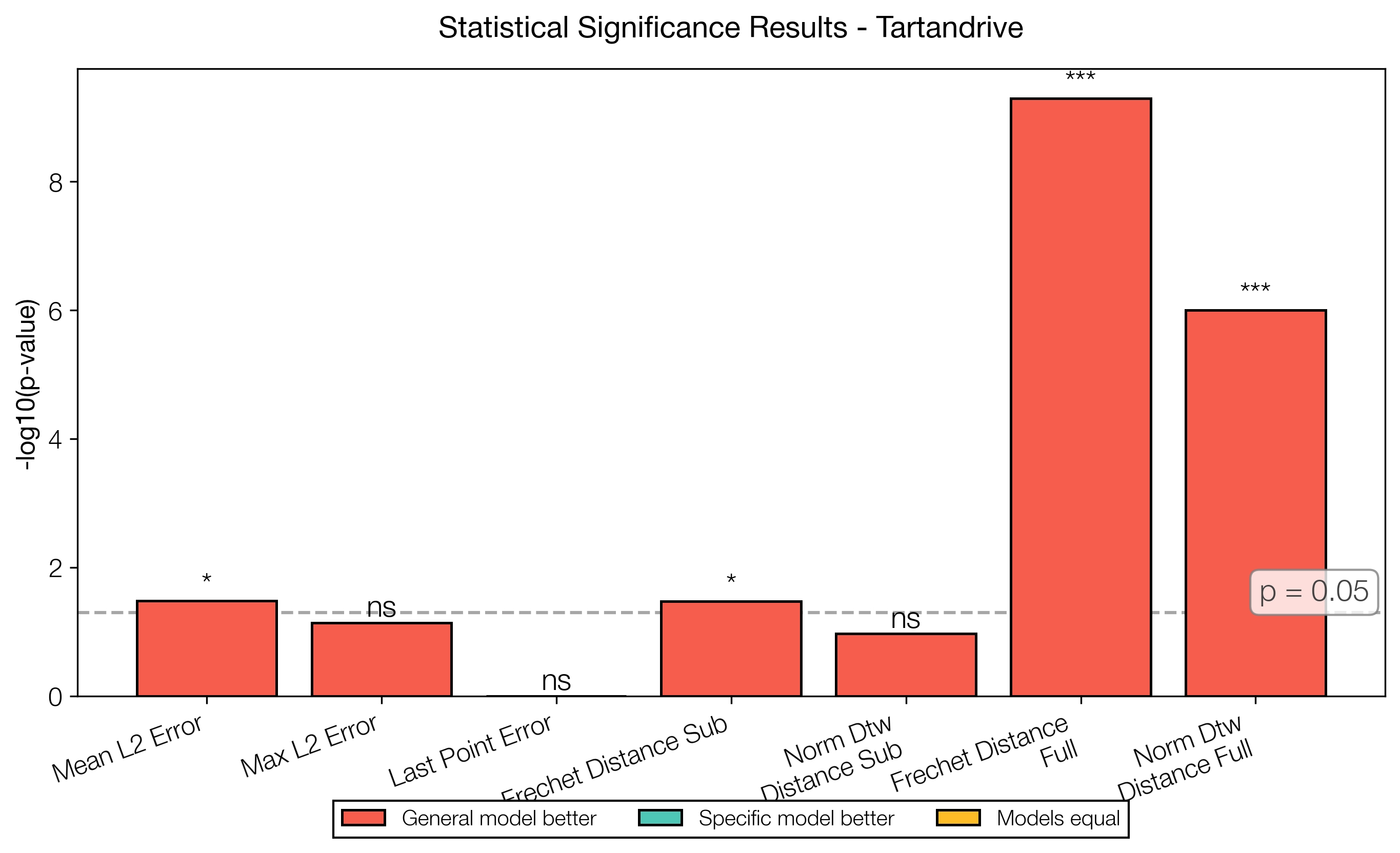}
    \caption{TartanDrive}
    \label{fig:tartandrive_significance}
  \end{subfigure}
  \hfill
  \begin{subfigure}[t]{0.48\textwidth}
    \includegraphics[width=\linewidth,height=0.25\textheight,keepaspectratio]{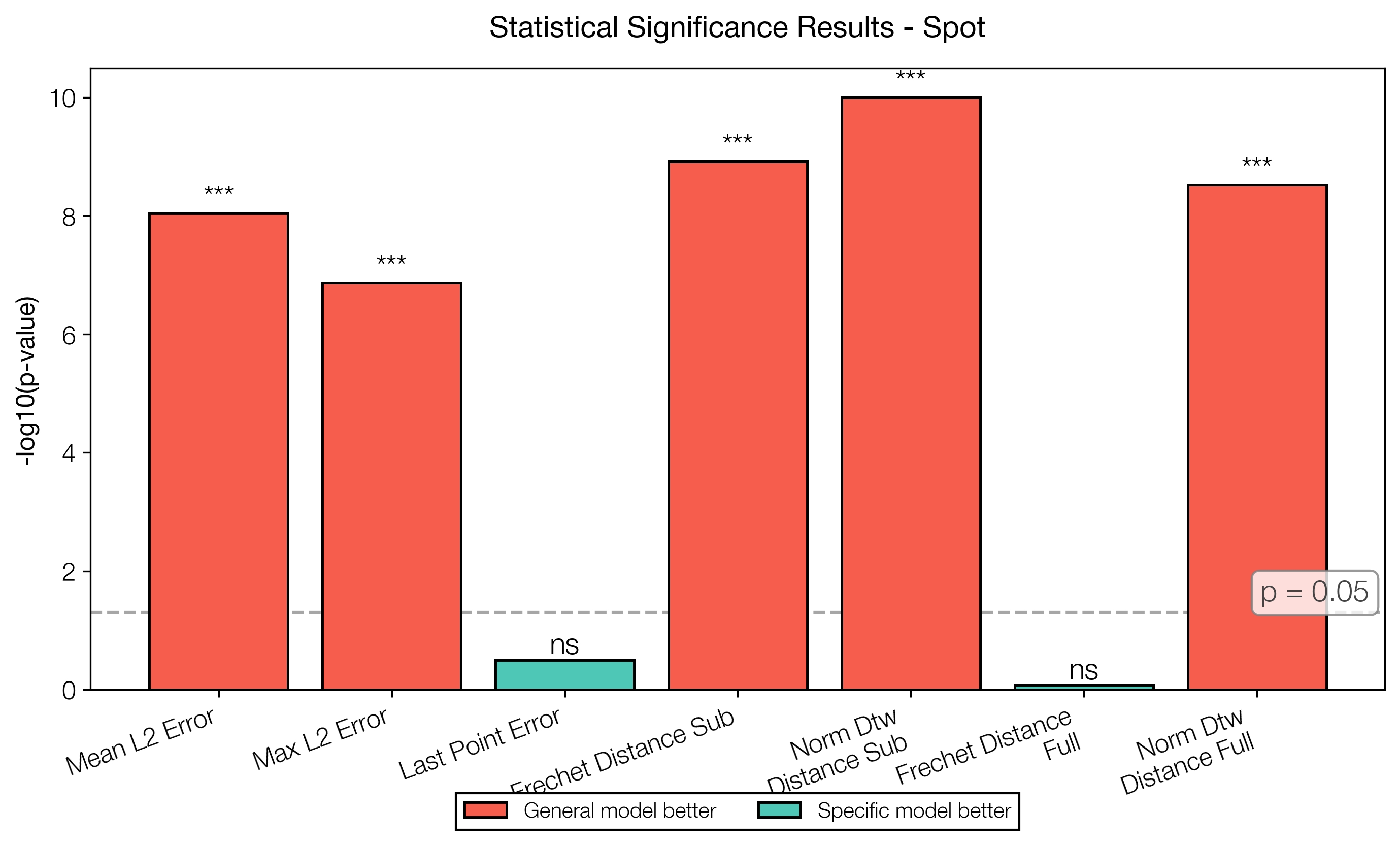}
    \caption{Spot}
    \label{fig:spot_significance}
  \end{subfigure}

  \caption{\textbf{Statistical significance of paired t-tests comparing generalist and robot-specific models} on four datasets (Tartandrive, CODA, SCAND, Spot) across seven trajectory-error metrics. Bars show $-\log_{10}(\text{p-value})$; green indicates the general model is better, orange the specific model, and gray cases where means are equal. `*' means $p < 0.05$, `**' means $p < 0.01$, `***' means $p < 0.001$, and "ns" means $p \geq 0.05$; the dashed line marks $p = 0.05$ ($-\log_{10}(0.05) \simeq 1.3$).}
  \label{fig:other_offline_metrics}
\end{figure*}

\end{document}